%% file: main.tex
\DocumentMetadata{} %
\documentclass[sigconf]{acmart} %

\usepackage{booktabs} %
\usepackage{float}
\restylefloat{table}
\usepackage{enumitem}
\usepackage{multirow}
\usepackage{colortbl}

\usepackage[most]{tcolorbox}

\theoremstyle{definition}
\newtheorem{definition}{Definition}[subsection]
\newtheorem{procedure}{}[subsection]
\newtheorem{discussion}{}

\newcommand{\gene}[1]{\textcolor{cyan}{#1}}
\newcommand{\anatomy}[1]{\textcolor{magenta}{#1}}
\newcommand{\disease}[1]{\textcolor{orange}{#1}}

\newcommand{\sre}{\texttt{\textbf{S}ubgraph \textbf{R}elevancy \textbf{E}stimation} }

\newcommand{\srank}{\texttt{\textbf{S}ubgraph \textbf{R}anker} }

\newcommand{\kbcd}{\texttt{\textbf{K}nowledge-Based \textbf{C}ausal \textbf{D}iscovery with Subgraphs as Prompt} }

\newcommand{\code}[1]{{\small\ttfamily#1}}

\copyrightyear{2025}
\acmYear{2025}
\setcopyright{cc}
\setcctype{by}
\acmConference[KDD '25]{Proceedings of the 31st ACM SIGKDD Conference on
Knowledge Discovery and Data Mining V.2}{August 3--7, 2025}{Toronto, ON,
Canada}
\acmBooktitle{Proceedings of the 31st ACM SIGKDD Conference on Knowledge
Discovery and Data Mining V.2 (KDD '25), August 3--7, 2025, Toronto, ON,
Canada}
\acmDOI{10.1145/3711896.3737076}
\acmISBN{979-8-4007-1454-2/2025/08}

\begin{document}

\title{Paths to Causality: Finding Informative Subgraphs Within Knowledge Graphs for Knowledge-Based Causal Discovery}

\renewcommand{\shorttitle}{Paths to Causality: Finding Informative Subgraphs Within KGs for Knowledge-Based Causal Discovery}

\author{Yuni Susanti}
\orcid{0009-0001-1314-0286}
\affiliation{%
  \institution{Fujitsu Limited}
  \city{Kawasaki}
  \country{Japan}
}
\email{susanti.yuni@fujitsu.com}

\author{Michael Färber}
\orcid{0000-0001-5458-8645}
\affiliation{%
  \institution{ScaDS.AI, TU Dresden} 
     \city{Dresden}
      \country{Germany}
  }  
\email{michael.faerber@tu-dresden.de}

\renewcommand{\shortauthors}{Yuni Susanti and Michael Färber}

\begin{abstract} 
Inferring causal relationships between variable pairs is crucial for understanding multivariate interactions in complex systems. \textit{Knowledge-based causal discovery}—which involves inferring causal relationships by reasoning over the \textit{metadata} of variables (e.g., names or textual context)—offers a compelling alternative to traditional methods that rely on observational data. However, existing methods using Large Language Models (LLMs) often produce unstable and inconsistent results, compromising their reliability for causal inference. To address this, we introduce a novel approach that integrates Knowledge Graphs (KGs) with LLMs to enhance knowledge-based causal discovery. Our approach identifies informative \textit{metapath}-based subgraphs within KGs and further refines the selection of these subgraphs using \textit{Learning-to-Rank}-based models. 
The top-ranked subgraphs are then incorporated into \textit{zero-shot} prompts, improving the effectiveness of LLMs in inferring the causal relationship. Extensive experiments on biomedical and open-domain datasets demonstrate that our method outperforms most baselines by up to 44.4 points in F1 scores,  evaluated across diverse LLMs and KGs. Our code and datasets are available on GitHub.\footnote{\url{https://github.com/susantiyuni/path-to-causality}}

\end{abstract}



\maketitle

\input{1-intro-rel-method}
\input{2-exp-disc-conc}

\input{3-appendix}

\bibliographystyle{ACM-Reference-Format}
\balance
\bibliography{ref} 

\clearpage

\end{document}

%% file: 1-intro-rel-method.tex
\section{Introduction}
\label{sec:intro}
Many of the most critical research questions involve inferring causal relationships~\cite{10.1162/tacl_a_00511}. For example, a physician must understand the potential side effects of a drug on a patient, as well as how it affects disease progression, before recommending a specific drug or treatment. 
Thus, inferring causal relationships between pairs of variables, such as \code{(smoking, lung cancer)}, is a fundamental step in causal discovery, as it enhances our understanding of the complex interactions among interconnected variables. 
Consequently, uncovering causal information through the analysis of observational data—a process known as \textit{causal discovery}—has recently garnered significant attention~\cite{Spirtes2001}.

Conventionally, causal discovery involves learning causal relations from observational data by using statistical methods and algorithms to infer causal structures and dependencies between variables, such as with PC and FCI algorithms~\cite{Spirtes2000}. %
Recently, Large Language Models (LLMs) offer a new perspective to tackle the causal discovery problem by reasoning on the \textit{metadata} associated with variables (e.g., variable names) instead of their actual data values--an approach referred to as \textit{knowledge-based causal discovery}~\cite{kic2023causal,SusantiKGSP}. 
In addition to being less dependent on large observational data, LLMs offer extensive prior knowledge drawn from diverse sources, such as scientific literature, databases, and domain-specific texts. This provides a rich context that observational data alone cannot offer. 
Thus, we argue that LLMs possess the capability to reveal complex causal patterns that traditional statistical methods might overlook, particularly in high-dimensional observational data. %

However, despite their strengths, LLMs often produce unstable results and may produce \textit{hallucinations} during reasoning, leading to incorrect conclusions and reducing their performance and reliability~\cite{hong2023faithfulquestionansweringmontecarlo,wang2023knowledgedrivencotexploringfaithful}. These issues significantly compromise the reliability of LLMs, especially for causal reasoning in sensitive domains such as healthcare and biomedicine. To address these issues, knowledge graphs (KGs) with their rich and diverse structured information have been integrated to enhance LLMs' reasoning, especially for question answering (QA), which needs both textual understanding and extensive real-world knowledge~\cite{Pan_2024}. 
KGs such as Wikidata~\cite{wikidata10.1145/2629489} (cross-domain) and \texttt{Hetionet}~\cite{hetionet10.7554/eLife.26726} (domain-specific) serve as reliable, interpretable sources for causal reasoning by capturing interconnections between information. 

\begin{figure}
  \centering
  \includegraphics[width=0.4\textwidth]{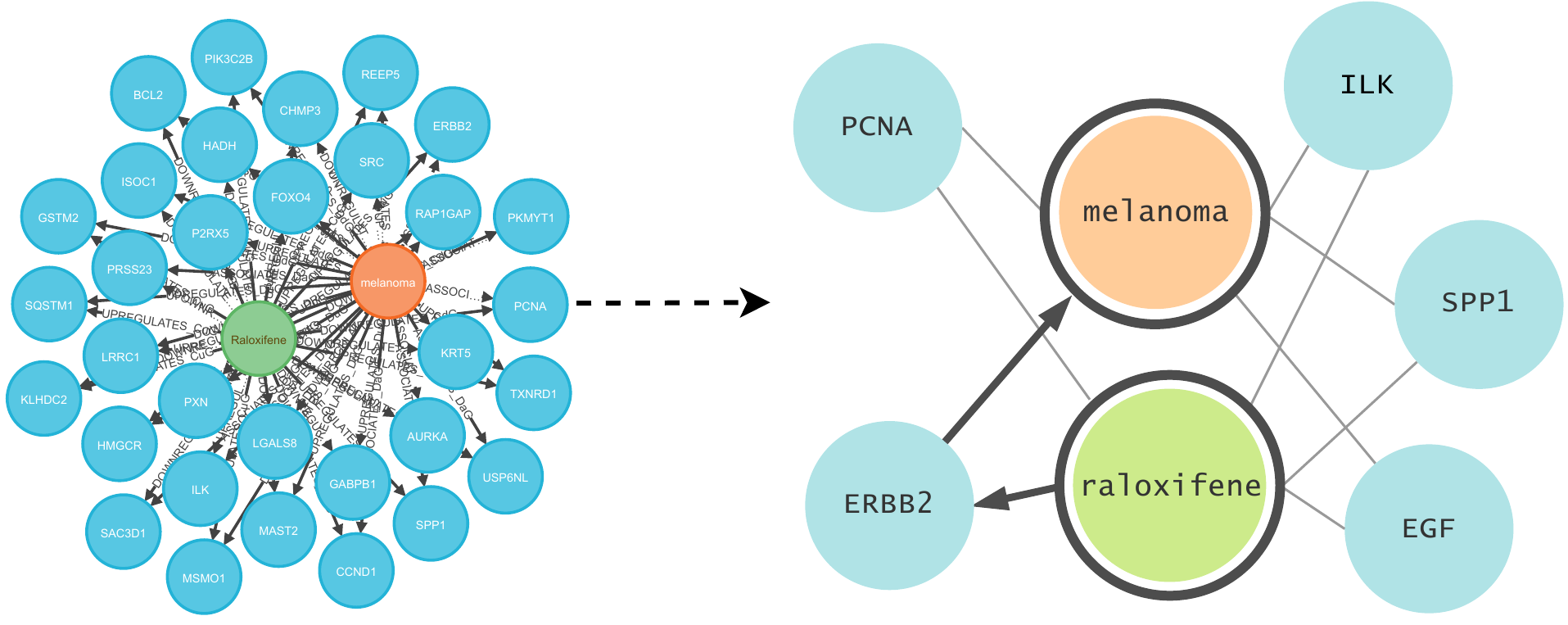}
    \caption{1-hop connections between \code{(raloxifene, melanoma)} in Hetionet~\cite{hetionet10.7554/eLife.26726}. Relation types are omitted for readability.} %
  \label{fig:multipaths}
  \Description{This image shows 1-hop connections between raloxifene and melanoma in Hetionet KG}
\end{figure}

The information in a KG that is relevant for causal reasoning is often complex, involving non-linear dependencies or associations that can be indirectly found through multi-hop paths. For example, in Figure~\ref{fig:multipaths}, the \textit{subgraph} \code{[raloxifene-ERBB2-melanoma]} provides valuable insights for inferring the causal relationship between \code{raloxifene} and \code{melanoma}. This subgraph indicates that \code{raloxifene}, a type of drug, upregulates gene \code{ERBB2}, which has been associated with disease \code{melanoma}.
This suggests the potential of \code{melanoma} as a side effect of taking \code{raloxifene}. A challenge lies in identifying such informative subgraphs within KGs for 
a given variable pair 
and using them with LLMs. As shown in Figure~\ref{fig:multipaths}, there are a total of 32 1-hop connections between \code{raloxifene} and \code{melanoma} in the KG, which increases to 1,031 subgraphs when extended to 2-hop connections. Naively processing all of these subgraphs would lead to an excessive number of LLM queries. While related works often select contextually relevant information from KGs based on vector similarity~\cite{surgekang2023knowledgegraphaugmentedlanguagemodels,TODshen-etal-2023-retrieval,KAPINGbaek-etal-2023-knowledge-augmented}, we argue that relying solely on similarity-based approach for identifying informative subgraphs is not optimal for the more complex task of causal discovery.

In this paper, we present a novel approach for integrating informative subgraphs from KGs with LLMs to enhance knowledge-based causal discovery. Our method focuses on identifying and leveraging informative \textit{metapath}-based subgraphs within KGs. First, an LLM estimates the relevance of the subgraphs in KGs with respect to determining causal relationships between variable pairs. Then, we develop specialized subgraph ranking models to refine the subgraph selection process, exploring various Learning-to-Rank approaches (\textit{pairwise, pointwise, listwise}) and algorithms (\textit{neural network} and classical \textit{gradient boosting} methods). This marks the first dedicated approach for identifying informative subgraphs within KGs for this task. Finally, we integrate the top-ranked subgraphs into a comprehensive framework for knowledge-based causal discovery. The framework incorporates the target variable pair, instructions, textual context, and top-ranked informative subgraphs into the prompt, enabling zero-shot reasoning by leveraging the pretrained knowledge of LLMs—without requiring large observational datasets or extensive supervised training.
Our experiments across multiple benchmark datasets from different domains show significant performance improvements of LLMs in inferring causal relationships between variable pairs, as evaluated across diverse language models and knowledge graphs. 

Overall, we make the following contributions:
\begin{itemize}[leftmargin=0.5cm]
\setlength{\itemsep}{3pt}
    \item We propose a new approach to enhance LLMs' causal reasoning by integrating them with KGs for zero-shot knowledge-based causal discovery, eliminating reliance on observational data. 
    \item We introduce specialized subgraph ranking models based on various Learning-To-Rank methods to refine subgraph selection from KGs, enhancing the performance of LLMs for inferring causal relationships by up to a 44.8-point increase in F1 score.
    \item We conduct extensive experiments on publicly available biomedical and open-domain datasets, demonstrating the effectiveness of our approach. Our method surpasses most baselines and shows generalizability across different LLMs and KGs.
    \item We compared our approach to conventional statistical causal discovery methods using observational data, achieving up to a 17-point reduction in Hamming distance and a 25.17-point F1 score improvement. This shows that LLMs, when integrated with KGs, effectively leverage semantic and contextual information of the variable, outperforming purely statistical methods.
\end{itemize} 

\section{Related Work}
\label{relwork}
\hspace{\parindent}\textbf{Knowledge-Based Causal Discovery.} Unlike \textit{statistical}-based causal discovery which uses purely observational data, \textit{knowledge-based causal discovery} focuses on the metadata associated with variables instead of their data values~\cite{kic2023causal,SusantiKGSP}.
This metadata may include variable names or any textual description related to the variables. Traditionally, such metadata-based causal reasoning relied on Subject Matter Experts (SMEs), but LLMs are now capable of providing knowledge that previously required SME expertise.
Recent works~\cite{kic2023causal,tu2023causaldiscovery,willig2022foundation,zhang2023understanding,gao-etal-2023-chatgpt} show that LLMs effectively provide background knowledge for causal discovery, notably,~\cite{kic2023causal} explores causal capabilities of LLMs by experimenting on cause-effect pairs. Their finding suggests that LLM-based prompting methods achieved superior performance than non-LLM approaches. Other recent studies have explored the use of LLMs to extract prior knowledge to be used for statistical-based causal discovery methods~\cite{takayama2024integratinglargelanguagemodels,ban2023querytoolscausalarchitects,2024iclrcausal}, however, it mainly relies on observational data. Other works focus on evaluating LLMs' ability to identify causal relations described in text~\cite{khetan-etal-2022-mimicause,SusantiKGSP,SusantiCEG,chatwal-etal-2025-enhancing},
focusing on identifying causal relationships rather than discovering new causal relations, which aligns with our work. Unlike previous research that relies on prompt engineering~\cite{kic2023causal,tu2023causaldiscovery,zhang2023understanding,gao-etal-2023-chatgpt,jiralerspong2024efficientcausalgraphdiscovery,chatwal-etal-2025-enhancing}, we enhance LLMs' capabilities for causal discovery by integrating \textit{informative} knowledge from KGs.

\textbf{KGs and LLMs.} 
Knowledge graphs (KGs) provide a reliable foundation for grounding LLMs' reasoning by organizing relational information and contextual knowledge in a structured manner. This enables retrieval-augmented generation (RAG) techniques, which effectively incorporate external knowledge, and is widely used in knowledge-intensive NLP tasks~\cite{rag10.5555/3495724.3496517} such as dialogue generation and question answering~\cite{hykgehypothesisknowledgegraph,knowledgptenhancinglargelanguage,KALMVbaek-etal-2023-knowledge-augmented-language,luo2024rog,surgekang2023knowledgegraphaugmentedlanguagemodels,TODshen-etal-2023-retrieval,KAPINGbaek-etal-2023-knowledge-augmented}. 

Research on integrating KGs with LLMs can be broadly categorized by the granularity of the retrieved information: \textit{entity}~\cite{fabula10.1145/3625007.3627505,TODshen-etal-2023-retrieval}, \textit{triple}~\cite{SusantiKGSP,KAPINGbaek-etal-2023-knowledge-augmented}, and \textit{subgraph}~\cite{surgekang2023knowledgegraphaugmentedlanguagemodels} levels. For instance, in the dialogue generation task, SURGE~\cite{surgekang2023knowledgegraphaugmentedlanguagemodels} augments LLMs by incorporating relevant subgraphs from KGs, which are extracted using a similarity-based approach combined with contrastive learning. %
Conversely, MK-TOD~\cite{TODshen-etal-2023-retrieval} operates at the entity level by identifying relevant KG entities based on vector similarities, 
also for dialogue generation. Similarly, KAPING~\cite{KAPINGbaek-etal-2023-knowledge-augmented} utilize embedding similarities to retrieve relevant triples from the KG for zero-shot QA.
\cite{SusantiKGSP} integrates KGs' structures (triples, subgraphs) into prompt-based learning for predicting causal relations; but it randomly selects information from KGs and relies on supervised learning to fine-tune the LLMs.
Given that KGs contain a vast amount of information, we focus on identifying relevant information to ensure that the model uses the most useful knowledge needed for knowledge-based causal discovery.

\begin{figure*}
  \centering
\includegraphics[width=0.98\textwidth]{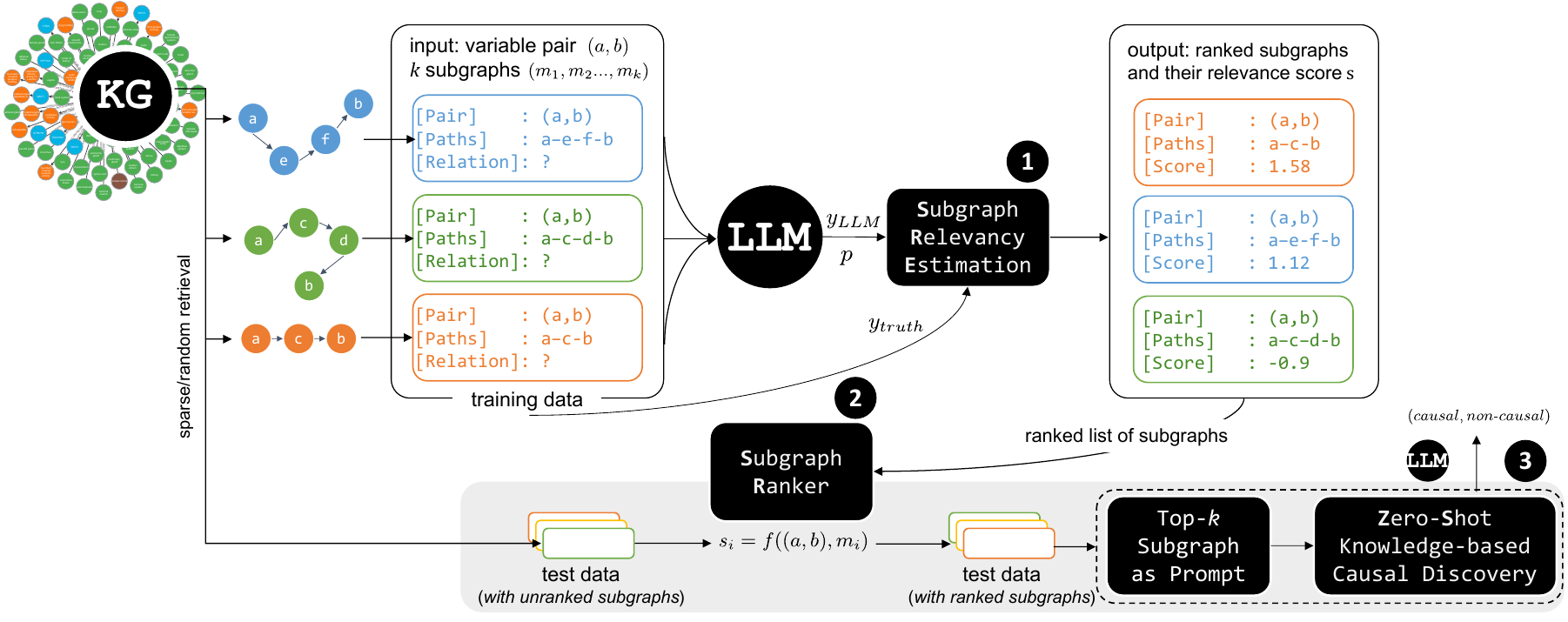}
\vspace{-2mm}
  \caption{Overview of the proposed approach, composed of three main modules: (1) \sre (2) \srank and (3) \kbcd(zero-shot). }
  \label{fig:framework}
    \Description{This image shows Overview of the proposed approach}
\end{figure*}

\begin{figure}
  \centering
\includegraphics[width=0.45\textwidth]{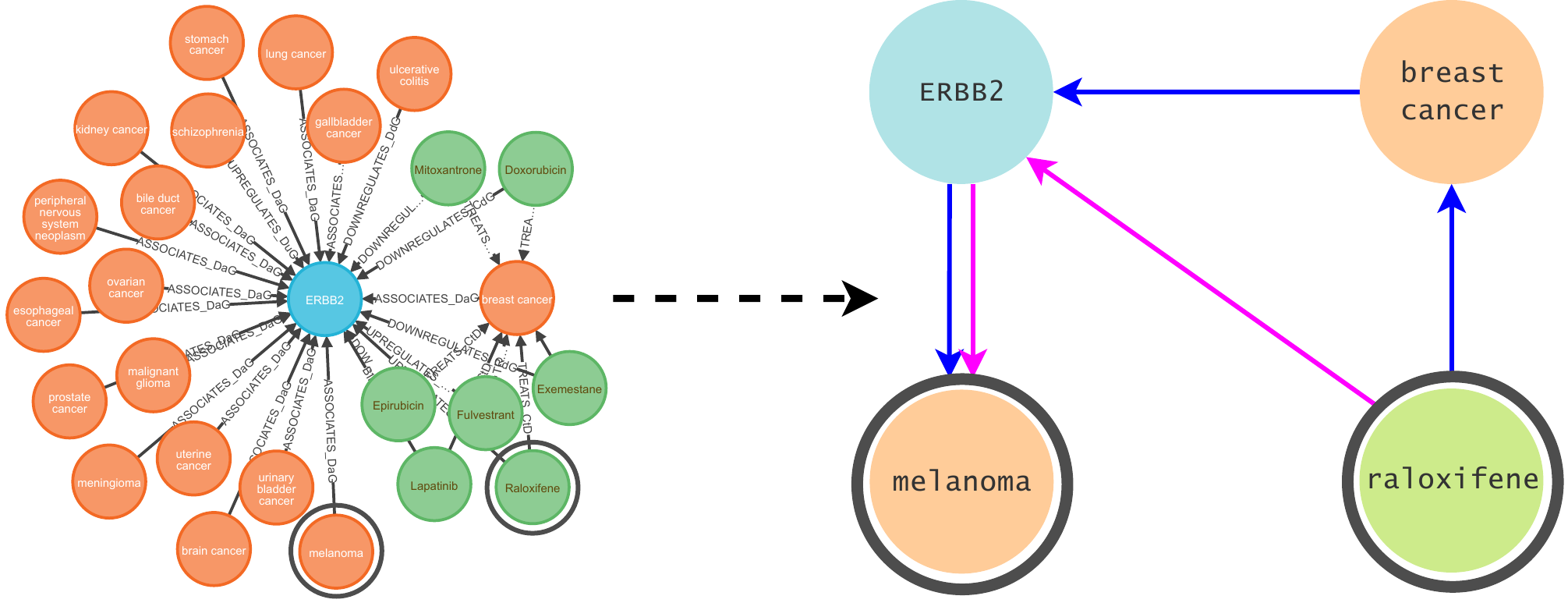}
\vspace{-3mm}
\caption{Metapaths connecting \code{raloxifene} and \code{melanoma}: (1) \code{[raloxifene-ERBB2-melanoma]} with \textcolor{magenta}{\code{[drug-gene-disease]}} node sequence, and (2) \code{[raloxifene-breast cancer-ERBB2-melanoma]} with \textcolor{blue}{\code{[drug-disease-gene-disease]}} node type sequence.}
  \label{fig:two_metapaths}
  \Description{This image shows Metapaths connecting raloxifene and melanoma}
\end{figure}

\section{Task Formulation}
\label{sec:task}
In this work, we focus on \textit{pairwise} knowledge-based causal discovery: given a pair of entities $a$ and $b$, i.e., variable or node pairs such as \code{(raloxifene, melanoma)}, the task is to predict if a causal relationship can be inferred between the pair. We formulate the task as \textit{classification task}, classifying the relation as \textit{causal} or \textit{non-causal}. We evaluate our approach on dataset $\mathcal{D} =\{\mathcal{X}, \mathcal{Y}\}$, where $\mathcal{X}$ is a set of data instance and $\mathcal{Y}=\{causal, non\text{-}causal\}$ is a set of relation labels. Each instance $x \in \mathcal{X}$ consists of a token sequence $x=\{w_{1}, w_{2},\ldots,w_{|n|}\}$ and the spans of a marked variable pair, and is annotated with a label $y_{x} \in \mathcal{Y}$. 

\section{Proposed Approach}
\label{sec:approach}
In a heterogeneous network like a KG, two nodes can be connected through many different paths. These paths--each being a \textit{subgraph} with a specific sequence of node types and relationships from an origin node to a destination node--are referred to as \textit{metapaths}~\cite{metapath10.14778/3402707.3402736}. 
Figure~\ref{fig:two_metapaths} illustrates two metapaths with different node type sequences connecting \code{raloxifene} and \code{melanoma}.

Such metapaths can capture various contextual and relational nuances, suggesting different underlying meanings, semantic interpretations, and potential causal implications. As a result, they are frequently used in explaining machine learning models~\cite{Chang_2024path,zhang2023pagelinkpathbasedgraphneural} and for network similarity analysis, especially in the biomedical domain~\cite{wangmeta1,yaometa2,Renmeta3}. In this domain, the term ``metapath'' itself refers to specific node type combinations thought to be \textit{informative} or meaningful~\cite{meta10.1093/bioinformatics/btad297}. Taking an example from Figure~\ref{fig:two_metapaths}, to determine the likelihood of a drug causing a specific disease, the metapath \code{[raloxifene-ERBB2-melanoma]} with \textcolor{magenta}{\code{[drug-gene-disease]}} node type sequence is more informative than \textcolor{blue}{\code{[drug-disease-gene-disease]}}.
The former suggests a direct biological mechanism where a \textcolor{magenta}{drug (\code{raloxifene})} influences (\code{upregulates}) a \textcolor{magenta}{gene (\code{ERBB2})}, which has been associated with a \textcolor{magenta}{disease (\code{melanoma})}, making it highly informative for inferring causality between the drug and the disease.

Inspired by this, we utilize the concept of metapaths to identify \textit{informative} subgraphs within KGs.
Given that many \textbf{\textit{metapath-based subgraphs}} may exist for a variable pair, the challenge is to pinpoint the most informative ones for determining causality. By utilizing these subgraphs, we aim to improve the accuracy and reliability of LLMs for knowledge-based causal discovery. %

Figure~\ref{fig:framework} illustrates the overview of our proposed approach, composed of the following three modules:
\begin{itemize}[leftmargin=0.5cm]
\setlength{\itemsep}{6pt}
    \item [1.] \sre utilizes LLMs to estimate the relevance of subgraphs, with the primary objective of generating a ranked list of subgraphs based on their informativeness for determining causal relations between variable pairs.
    \item[2.]\srank 
    distills the LLMs' capability in inferring causal relationship by training specialized subgraph ranker models using the generated ranked list of subgraphs. This module further refines the subgraph selection, ultimately selecting the most informative subgraph for inferring causality.
    \item[3.]\kbcd performs \textit{zero-shot} knowledge-based causal discovery on variable pairs given the top-ranked subgraphs ranked by \texttt{\textbf{S}ubgraph \textbf{R}anker} model, as a prompt for LLMs.
\end{itemize}

We elaborate the proposed approach in the following subsections. We start with preliminaries (\S\ref{sec:preliminaries}), followed by the detailed explanation of \sre (\S\ref{sec:sre}) and \srank (\S\ref{sec:reranker}) modules. Lastly, we explain the integration of informative subgraphs from KGs with LLMs in \kbcd (\S\ref{sec:kbcd}).

\subsection{Preliminaries}
\label{sec:preliminaries}
\begin{definition}{\textbf{Subgraph.}}
Let $\mathcal{G} = (V, E)$ be a graph with a vertex set $V$ and an edge set $E$. 
A subgraph $\mathcal{G'}=(V', E')$ of a graph $\mathcal{G}$ is then another graph whose vertex set and edge set are subsets of those of $\mathcal{G}$, i.e., where $V' \subseteq V$ and $E' \subseteq E$.
\end{definition}

\begin{definition}{\textbf{Knowledge Graph.}}
    A knowledge graph is a specific type of graph representing a network of entities and the relationships between them. Formally, we define a knowledge graph as a directed labeled graph $\mathcal{KG} = (N, E, R, \mathcal{F})$, where $N$ is a set of nodes (entities), $E \subseteq N \times N$ is a set of edges (relations), $R$ is a set of relation labels, and $\mathcal{F}: E \to R$, is a function assigning edges to relation labels. For instance, assignment label $r$ to an edge $e=(x,y)$ can be viewed as a triple $(x, r, y)$, e.g., \code{(Tokyo, IsCapitalOf, Japan)}. 
\end{definition}

\begin{definition}{\textbf{Metapath-based Subgraph.}}
    In this work, we focus on \textit{metapath}-based subgraphs within a KG. 
    Formally, a metapath-based subgraph $\mathcal{M}$ can be defined as a path that is denoted in the form of $T_{1}\xrightarrow{\mathit{R_{1}}}T_{2}\xrightarrow{\mathit{R_{2}}} \ldots \xrightarrow{\mathit{R_{n}}}T_{n+1}$ describing a composite relation $R = R_1\circ R_2\circ \ldots \circ R_{n} $ between node types $T_{1}$ and $T_{n+1}$\cite{metapath10.14778/3402707.3402736}. 
This illustrates how two nodes are interconnected through a series of relationships and sequences of node types, providing context, i.e., semantics, and potential causal meaning within a KG. Note that when we refer to a \textbf{subgraph} in this paper, we specifically mean a \textbf{metapath-based subgraph}. We also use the terms \textbf{variable} and \textbf{entity} pairs interchangeably throughout the paper.

\end{definition}
\subsection{\textbf{S}ubgraph \textbf{R}elevancy \textbf{E}stimation}
\label{sec:sre}
In this module, we use LLMs to estimate how \textit{informative} or \textit{relevant} a subgraph is for inferring causality. The goal is to generate a ranked list of subgraphs based on their relevance for accurately determining causal connections between variable pairs. Formally, given a pair \((a, b)\) and \(k\) subgraphs \((m_1, m_2, \ldots, m_k)\) extracted from a KG, \(\sre\) produces a ranking \((r_1, r_2, \ldots, r_k)\), where \(r_i \in \{1, 2, \ldots, k\}\) denotes the rank of subgraph \(m_i\).


As illustrated in Figure~\ref{fig:framework} (1), given the variable pair and a set of $k$ subgraphs\footnote{We employ \textit{sparse/random retrieval} to filter the subgraphs (see Appendix \ref{app:kg}.} as input prompt to the LLM, we  instruct the LLM to predict the causal relationship between the variable pair by outputting the relation, e.g., $(causal, non\text{-}causal)$. A subgraph is considered as relevant if the LLM predicts the relation correctly, according to the human-labeled ground truth relation. The relevance score $s_i$ for each subgraph is then calculated based on the probability $p$ of the LLM generating the prediction, as follows:
\begin{equation}
\label{eq:relscore}
s_{i} = \begin{cases}1+p('causal/non\text{-}causal') & if \quad y_{LLM} = y_{truth}\\1-p('causal/non\text{-}causal') & if \quad y_{LLM} \neq y_{truth}\end{cases}
\end{equation}
where $p('causal/non\text{-}causal')$ denotes the log probability of LLMs generating the prediction, $y_{LLM}$ denotes the output of an LLM, and $y_{truth}$ denotes the ground truth of the relation. 
Finally, we finalize the ranking of the subgraphs: given the set of subgraphs \( M = \{ m_1, m_2,\ldots, m_k \} \) with scores \( \{ s_1, s_2,\ldots,s_k \} \), we assign the ranking $r_i = \text{argsort}_{i}(s_{1},s_{2}...,s_{n})$ for the subgraphs.
The final output is a dataset $\mathcal{R}$ containing variable pairs with subgraphs ranked based on their \textit{informativeness} in inferring the causal relation. We provide examples of prompts and ranked lists of subgraphs
in Appendix \ref{app:sre}. %

\subsection{Subgraph Ranker}
\label{sec:reranker}
Ranking plays a crucial role in our approach, as KGs often include multiple subgraph connections between a variable pair. 
In our approach, we introduce specialized models for subgraph ranking to further refine the subgraph selection process, utilizing additional \textit{meta} information from the metapath-based subgraphs, e.g., node/edge type sequences. 
As illustrated in Figure~\ref{fig:framework} (2), we train a dedicated subgraph ranker using the 
ranked list of subgraphs produced by the \sre module, thereby \textit{distilling} the LLM's capability to infer causal relationships. 

Ranking problems are prevalent in Information Retrieval field, including tasks like web search ranking and text retrieval. In these tasks, given a query $q$ and a relevant list of documents $D$, 
\textit{Learning-to-Rank} (LTR) methods learn a function to predict the relevance scores of the documents based on the given query, arranging the documents in an ordered list~\cite{IR10.1561/1500000016}. 
Inspired by this, we formulate the subgraph ranking as an Information Retrieval task. Our proposed \srank takes a variable pair $(a,b)$ and the set of subgraphs $(m_{1},m_{2}\dots,m_{k})$ to train an LTR-based ranker model. This model ranks the subgraphs based on their \textit{informativeness} in inferring the causal relationship between the variables $(a,b)$.

\subsubsection{\emph{\textbf{Training Objective}}}
Formally, \srank is a ranking model $s_{i} = f_{\theta}((a,b),m_{i})$ with parameter $\theta$ which computes a relevance score \(s_i\) for input pair \(((a,b), m_i)\). 
Let $M = \{m_1, m_2, \ldots, m_k\}$ be a set of subgraphs, with their corresponding predicted scores $\{s_1, s_2, \ldots, s_k\}$ and ground truth relevance scores $\{y_1, y_2, \ldots, y_k\}$. Using the ranked dataset $\mathcal{R}$ generated by \sre (\S\ref{sec:sre}), we consider the following \textbf{\textit{pointwise}}, \textbf{\textit{pairwise}} and \textbf{\textit{listwise}} objective functions to optimize the ranker model.

\paragraph{\textbf{Pointwise}} In the pointwise approach, the subgraph ranking is formulated as a \textit{regression} task. That is, each subgraph is scored independently and the loss function minimizes the difference between the predicted and ground truth relevance scores. We consider a loss function based on \textit{Root Mean Square Error} (RMSE):
\begin{equation}
\mathcal{L}_{\text{RMSE}} = \sqrt{\frac{1}{n} \sum_{i=1}^k (s_i - y_i)^2}
\end{equation}

\paragraph{\textbf{Pairwise}} In the pairwise approach, we consider a loss function based on RankNet~\cite{Burges2010FromRT}, defined as follows:
\begin{equation}
    \mathcal{L}_{\text{RankNet}} = \sum_{i=1}^{k} \sum_{j=1}^{k} \mathbb{I}_{r_i < r_j} \log (1 + \exp (s_i - s_j)) 
\end{equation}
RankNet is a pairwise loss that measures the correctness of relative subgraph orders, thus, for a variable pair with $k$ subgraphs, we can construct in total $k (k-1) / 2$ pairs. 

\paragraph{\textbf{Listwise}} In the listwise approach, we consider the ListNet~\cite{Listnetcao2007learning} objective function. We use ListNet softmax loss as defined in Eq.~\ref{eq:listnet}, implemented in~\cite{AllRankPobrotyn2020AwareLT}. 
Instead of modeling the probability of a pairwise comparison using scoring difference, ListNet models the probability of the entire ranking result.
\begin{equation}
\label{eq:listnet}
\mathcal{L}_{\text{ListNet}} = -\sum_{i=1}^{k} \text{softmax}(y)_i \times \log( \text{softmax}(s)_i)
\end{equation}
\begin{figure}
  \centering
  \includegraphics[width=0.28\textwidth]{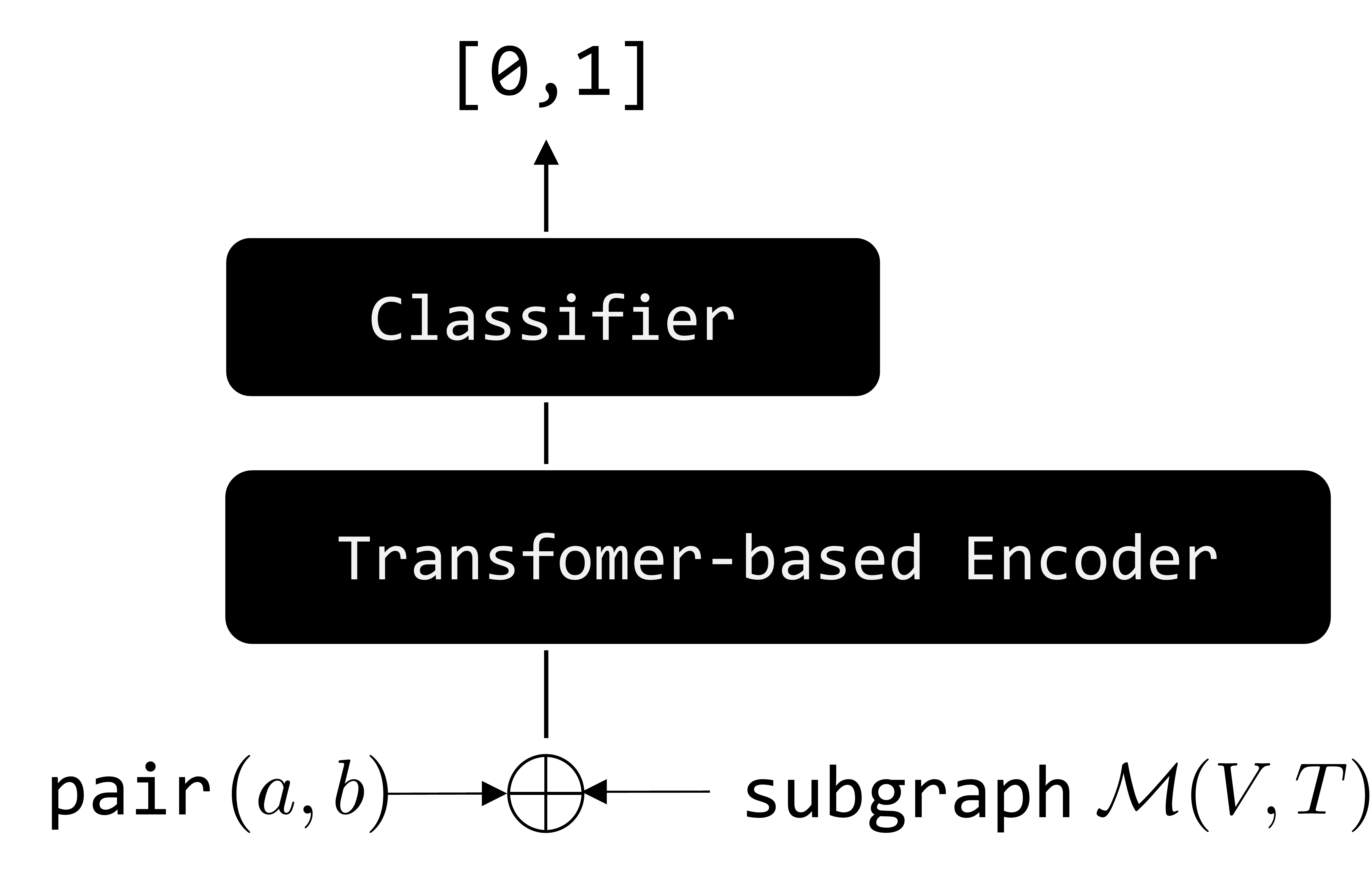}
  \vspace{-3mm}
  \caption{\srank cross encoder model architecture with Transformer-based encoder.}
  \label{fig:ce}
  \Description{This image shows cross encoder model architecture with Transformer-based encoder}
\end{figure}

\subsubsection{\emph{\textbf{Model Architecture}}} 
As shown in Figure~\ref{fig:ce}, our \srank is built on a \code{Transformer}-based cross-encoder, a model widely used for text ranking~\citep{Nogueira2019PassageRW,bertrank10.1007/978-3-030-72240-1_26}. A cross-encoder processes two text inputs simultaneously, encoding them into a single representation and generating a relevance score. We adapt this approach by concatenating variable pair $(a,b)$ with subgraph $\mathcal{M}$, as follows:
\begin{equation}
\label{eq:concat}
\begin{aligned}
(a, b) \oplus \mathcal{M}\{V, T\} = [\texttt{CLS}] a, b [\texttt{SEP}] (t_1 v_1, t_2 v_2, \ldots, t_n v_n)
\end{aligned}
\end{equation}
where $\mathcal{M}\{V,T\}$ is a subgraph $\mathcal{M}$ with path length $n$ containing a set of nodes 
$V=(v_{1}, v_{2}, \ldots v_{|n|})$
with their corresponding node types $T=(t_{1}, t_{2}, \ldots t_{|n|})$.
We use the \textit{meta} information \textit{node type} for training the ranker, as it is considered informative~\cite{metapath10.14778/3402707.3402736}. The node $v_1$ corresponds to $a$, and the last node $v_n$ corresponds to $b$. We insert special tokens \code{[CLS]} and \code{[SEP]} as shown in Eq.~\ref{eq:concat} following the practice of the Transfomer-based encoder~\cite{BERTDBLP:journals/corr/abs-1810-04805}. Lastly, we use the representation of \code{[CLS]} to estimate the relevance score with a classification layer, as illustrated in Figure~\ref{fig:ce}. 

Studies have shown that classical Learning-to-Rank methods, especially Gradient Boosting Decision Trees (GBDT), often outperform neural-based approaches, including cross encoder~\cite{50030}. Thus, we train a GDBT-based subgraph ranker as an alternative. We concatenate the pair with the subgraph as in Eq.~\ref{eq:concat} and train an \textit{n-gram} model on these combined sequences. This model is then used to extract features for training the gradient boosting-based subgraph ranker. Details of this approach are provided in Appendix \ref{app:xgboost}. %

\subsection{Knowledge-Based Causal Discovery with Subgraphs as Prompt}
\label{sec:kbcd}
In this module, we perform the knowledge-based causal discovery task. For a given variable pair $(a, b)$, we prompt an LLM to predict if a causal relationship can be inferred between them. The prompt also includes textual context and the top-$k$ subgraphs, as ranked by the \srank model.
The following describes the details:

\begin{figure}
  \centering
  \includegraphics[width=0.33\textwidth]{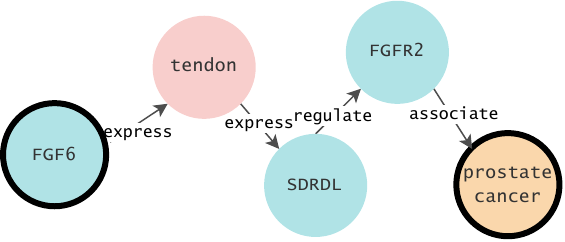}
  \vspace{-3mm}
  \caption{Subgraph example. Different node colors represent different node types: \gene{gene}, \anatomy{anatomy}, \disease{disease}}
  \label{fig:subgraph_as_prompt}
    \Description{This image shows Subgraph example.}
\end{figure}

\begin{procedure}\textbf{Selecting Top-$k$ Subgraphs.}
    Given the ranked subgraphs produced by \srank, we select the top-$k$ most relevant subgraphs.
    Let $\{(m_1, s_1), (m_2, s_2), \ldots, (m_n, s_n)\}$ be the set of subgraphs and their scores for a pair $(a,b)$. The top-$k$ subgraphs are selected as follows:
    \begin{equation}
    \text{top-k}((a, b)) = \{m_i \mid i \in \text{argsort}(s_1, s_2, \ldots, s_n)[:k]\}
    \end{equation}
    where $\text{argsort}(s_1, s_2, \ldots, s_n)$ is a function that returns the indices of the sorted scores
    and $[:k]$ selects the top-k indices from the list.
\end{procedure}

\begin{procedure}{\textbf{Subgraphs as Prompt.}}
    We further transform the top-\textit{k} subgraphs into prompts by converting the structures into a sequence of elements that can be processed sequentially. 
    We \textit{verbalize} the subgraphs by converting the sequential relationships between nodes into natural language text, as follows:
    \begin{equation}
        \label{eq:metapath}
        \begin{aligned}
        &\mathcal{M}\{V,E,T\}=\text{\code{``Relation paths between the pair: ''}} \\
        &\:\: \{(t_1 v_1, e_1, t_2 v_2), (t_2 v_2, e_2, t_3 v_3), \ldots, (t_{n-1} v_{n-1}, e_n, t_n v_n)\}
        \end{aligned}
    \end{equation}
    where $\mathcal{M}\{V,E,T\}$ is a subgraph $\mathcal{M}$ with path length $n$ containing a set of nodes 
    $V=(v_{1}, v_{2}, \ldots v_{|n|})$
    with their corresponding node types $T=(t_{1}, t_{2}, \ldots t_{|n|})$
    Each edge \(e_i\) connects node \(v_i\) to its immediate neighbor \(v_{i+1}\) within the subgraph. 
    The string \code{``Relation paths...''} are optional and can be replaced by other tokens. 

Consider a subgraph as illustrated in Figure~\ref{fig:subgraph_as_prompt}. With Eq.~\ref{eq:metapath}, the subgraph can be linearized into a prompt, as follows:

\begin{tcolorbox}[boxsep=2pt,left=2pt,right=2pt,top=2pt,bottom=2pt]
\code{\gene{Gene} FGF6 $\xrightarrow{express}$ \anatomy{anatomy} tendon, \anatomy{anatomy} tendon $\xrightarrow{express}$ \gene{gene} SDRDL, \gene{gene} SDRDL $\xrightarrow{regulate}$ \gene{gene} FGFR2, \gene{gene} FGFR2 $\xrightarrow{associate}$ \disease{disease} prostate cancer}
\end{tcolorbox}

This representation captures the sequential relationships between entities and their corresponding relations in a linear format suitable for use in prompts. As it is technically a \textit{metapath}, it includes the \textit{meta} information (node types $T$), as well. Studies have shown that giving more textual context does not always lead to better performance, and sometimes a simpler prompt can be more beneficial than a complex one~\cite{PromptPetroniLPRWM020,Prompt10.1145/3411763.3451760}. Therefore, we can also opt to exclude the node types $T$ and relation labels $E$ from the prompt. In such case, the prompts would be as follows:

\begin{tcolorbox}[boxsep=2pt,left=2pt,right=2pt,top=2pt,bottom=2pt]
\code{FGF6 $\xrightarrow{express}$ tendon $\xrightarrow{express}$ SDRDL $\xrightarrow{regulate}$ FGFR2 $\xrightarrow{associate}$ prostate cancer}
\tcblower
\code{FGF6$\rightarrow$ tendon$\rightarrow$ SDRDL$\rightarrow$ FGFR2$\rightarrow$ prostate cancer}
\end{tcolorbox}

These representations capture the subgraph structure in its simplest form, while still suitable for processing by LLMs designed for sequential inputs. The arrow symbols \texttt{``$\rightarrow$''} are optional and can be replaced by other fitting symbols or tokens (e.g., hyphen \texttt{``--''}).
\end{procedure}

\begin{procedure}{\textbf{Zero-Shot Knowledge-Based Causal Discovery.}} In this final step, the LLM is tasked with inferring a causal relationship between a variable pair based on the provided information in the prompt. We perform knowledge-based causal discovery with no training examples or parameter updates to the LLMs, i.e., a \textit{zero-shot} approach. A zero-shot approach enables the model to predict causal relationships between unseen variable pairs without any training/fine-tuning to the model. In our work, this method is preferable given the scarcity of causal datasets. 

Formally, given $\mathcal{I}$ as the instruction, $x$ as the textual context, and $\mathcal{M}$ as the subgraph prompt, the final full prompt $x'$ as the input to the LLMs for the pair $(a,b)$ can be formulated as follows: 
\begin{equation}
\label{eq:mlm}
x' = [\mathcal{I}] [x] \: [\mathcal{M}] \: \text{\code{The relation between}}\:[a]\:\text{\code{and}}\:[b]\:\text{\code{is}}
\end{equation}
\end{procedure}

Additional tokens \code{``The relation between...''} are optional and can be replaced by other fitting templates. Additional information on this step is provided in Appendix \ref{app:kbcd}.

%% file: 2-exp-disc-conc.tex
\section{Evaluation}
\subsection{Experiment Settings} 
\label{sec:expset}
We evaluate the proposed approach in a knowledge-based causal discovery task: %
given a pair of entities $a$ and $b$, the task is to predict if a causal relationship can be inferred between the pair. We specifically focus on integrating informative subgraphs from KGs to enhance the internal knowledge of the LLMs in inferring causality.

To identify informative subgraphs in the KGs, we run \sre (\S\ref{sec:sre}) using the reliable and cost-efficient \texttt{mistral-7b}~\cite{jiang2023mistral7b} model, generating an up to $k$=10 ranked subgraphs dataset per variable pair (example in Appendix~\ref{app:sre}). Next, to further refine the subgraph selection, we train \srank (\S\ref{sec:reranker}) model using this dataset, experimenting with several cross encoders and \code{XGBoost}~\cite{xgb10.1145/2939672.2939785}-based models.
Due to comparable performance across cross encoders, (provided in Appendix~\ref{app:srank}), we selected (1) \code{roberta-base} cross encoder and (2) \code{XGBoost}-based models for ranking the subgraphs that would be used for the subsequent zero-shot knowledge-based causal discovery (\S\ref{sec:kbcd}) experiments.

Implementation details for each module of the proposed approach are in Appendices \ref{app:sre}, \ref{app:srank}, \ref{app:kbcd}, with baseline details in Appendix~\ref{app:baseline}. Computational complexity and hardware specifications are in Appendix~\ref{app:comp-hw}. Code and resources are available on GitHub.

\subsection{\textbf{Choice of LLMs and KGs}}
\label{sec:slmchoice} 
Recent studies have shown that 7B LLMs deliver relatively strong performance across various tasks~\cite{7b10434081,7bli2024common7blanguagemodels}. Thus, we experiment with the following widely-used 7--8B LLMs: 
\begin{itemize}[leftmargin=0.5cm]
    \item \texttt{mistral-7b-instruct}: a highly efficient model with high-level performance, surpassing even 16-34B models~\cite{jiang2023mistral7b}. 
    \item \texttt{llama-3-8b-instruct}: the 8B model from Meta's Llama series offering improved reasoning capabilities~\cite{llama3}.
    \item \texttt{gemma-7b-it}: a lightweight LLM based on Google’s Gemini~\cite{gemmateam2024gemmaopenmodelsbased}.
\end{itemize}

\noindent For the knowledge graphs, we select the following: 
\begin{itemize}[leftmargin=0.5cm]
    \item \texttt{Wikidata}~\cite{wikidata10.1145/2629489}, as the open-domain KG. 
    \item \texttt{Hetionet}~\cite{hetionet10.7554/eLife.26726}, a domain-specific KG assembled from 29 different databases, covering genes, compounds, and diseases.
\end{itemize}

We select \texttt{Wikidata} for its wide coverage of subjects and 
\texttt{Hetionet} since we primarily evaluate on biomedical domain datasets. Details on querying the KGs and subgraphs extraction are in Appendix~\ref{app:kg}. 

\subsection{Datasets}
\label{sec:data}

Since causality is often studied in the biomedical domain, we primarily evaluate our approach in this field using \code{GENEC}~\cite{SusantiKGSP} for gene-gene relationships, \code{ADE}~\cite{ADEGURULINGAPPA2012885} for drug-side effect relationships, and \code{COMAGC}~\cite{COMAGCLee2013com} for gene-disease relationships. Additionally, we evaluate it on the open-domain dataset \code{SEMEVAL}~\cite{SEMhendrickx-etal-2010-semeval}. 
The dataset includes the ground truth causal relationships labeled by human experts, allowing us to assess the models' prediction performance in the evaluation.
The dataset details and examples are in Appendix~\ref{app:dataset}.

\begin{table*}
\small
\centering
\vspace{-1mm}
\begin{tabular}{l|rrr|rrr|rrr}
\toprule
\multicolumn{1}{c|}{} & \multicolumn{3}{c|}{\texttt{COMAGC}~\cite{COMAGCLee2013com}} & \multicolumn{3}{c|}{\texttt{GENEC}~\cite{SusantiKGSP}} & \multicolumn{3}{c}{\texttt{ADE}~\cite{ADEGURULINGAPPA2012885}}  \\
\multicolumn{1}{c|}{} & \multicolumn{1}{c}{P} & \multicolumn{1}{c}{R} & \multicolumn{1}{c|}{F1} & \multicolumn{1}{c}{P} & \multicolumn{1}{c}{R} & \multicolumn{1}{c|}{F1} & \multicolumn{1}{c}{P} & \multicolumn{1}{c}{R} & \multicolumn{1}{c}{F1} \\

\midrule
\multicolumn{1}{c}{} & \multicolumn{9}{c}{\textcolor{blue}{\texttt{mistral-7b-instruct~\cite{jiang2023mistral7b}}}} \\
\midrule
\texttt{(baseline) no-subgraph} & 66.67 & 47.06 & 55.17 & 27.27 & 16.67 & 20.69 & 100.00 & 36.84 & 53.85 \\
\texttt{(baseline) random subgraph selection} & 90.00 & 52.94 & 66.67 & 100.00 & 16.67 & 28.57 & 100.00 & 52.63 & 68.97 \\
\texttt{(baseline) similarity-based ranker} & 76.92 & 58.82 & 66.67 & 100.00 & 16.67 & 28.57 & 100.00 & 55.26 & 71.19 \\
\texttt{(baseline) GPT-based ranker} & 93.33 & 82.35 & \underline{\textbf{87.50}} & 60.00 & 16.67 & 26.09 & 100.00 & 57.89 & \textbf{73.33} \\ 
\rowcolor[HTML]{ECF4FF}\texttt{(ours) Subgraph Ranker (rmse-cross encoder)} & 86.67 & 76.47 & 81.25 & 83.33 & 27.78 & \textbf{41.67} & 100.00 & 55.26 & 71.19 \\
\rowcolor[HTML]{ECF4FF}\texttt{(ours) Subgraph Ranker (ranknet-cross encoder)} & 85.71 & 70.59 & 77.42 & 83.33 & 27.78 & \textbf{41.67} & 100.00 & 63.16 & \underline{\textbf{77.42}} \\
\rowcolor[HTML]{ECF4FF}\texttt{(ours) Subgraph Ranker (listnet-cross encoder)} & 81.25 & 76.47 & 78.79 & 100.00 & 27.78 & \underline{\textbf{43.48}} & 100.00 & 57.89 & \textbf{73.33} \\
\rowcolor[HTML]{ECF4FF}\texttt{(ours) Subgraph Ranker (XGBoost)} & 92.86 & 76.47 & \textbf{83.87} & 80.00 & 22.22 & 34.78 & 100.00 & 57.89 & \textbf{73.33} \\
 \midrule
\multicolumn{1}{c}{} & \multicolumn{9}{c}{\textcolor{blue}{\texttt{llama-3-8b-instruct~\cite{llama3}}}} \\
\midrule
\texttt{(baseline) no-subgraph} & 57.14 & 23.53 & 33.33 & 66.67 & 44.44 & 53.33 & 100.00 & 50.00 & 66.67 \\
\texttt{(baseline) random subgraph selection} & 61.11 & 64.71 & 62.86 & 43.75 & 38.89 & 41.18 & 100.00 & 71.05 & 83.08 \\
\texttt{(baseline) similarity-based ranker} & 60.00 & 70.59 & 64.86 & 46.67 & 38.89 & 42.42 & 100.00 & 76.32 & 86.57 \\
\texttt{(baseline) GPT-based ranker} & 68.42 & 76.47 & \textbf{72.22} & 55.56 & 55.56 & \textbf{55.56} & 100.00 & 78.95 & 88.24 \\
\rowcolor[HTML]{ECF4FF}\rowcolor[HTML]{ECF4FF}\texttt{(ours) Subgraph Ranker (rmse-cross encoder)} & 63.64 & 82.35 & 71.79 & 50.00 & 38.89 & 43.75 & 100.00 & 76.32 & 86.57 \\
\rowcolor[HTML]{ECF4FF}\texttt{(ours) Subgraph Ranker (ranknet-cross encoder)} & 73.68 & 82.35 & \underline{\textbf{77.78}} & 58.82 & 55.56 & \underline{\textbf{57.14}} & 100.00 & 81.58 & \textbf{89.86} \\
\rowcolor[HTML]{ECF4FF}\texttt{(ours) Subgraph Ranker (listnet-cross encoder)} & 65.00 & 76.47 & 70.27 & 61.54 & 44.44 & 51.61 & 100.00 & 81.58 & \textbf{89.86} \\
\rowcolor[HTML]{ECF4FF}\texttt{(ours) Subgraph Ranker (XGBoost)} & 60.00 & 70.59 & 64.86 & 46.15 & 33.33 & 38.71 & 100.00 & 86.84 & \underline{\textbf{92.96}} \\
 \midrule
\multicolumn{1}{c}{} & \multicolumn{9}{c}{\textcolor{blue}{\texttt{gemma-7b-it~\cite{gemmateam2024gemmaopenmodelsbased}}}} \\
\midrule
\texttt{(baseline) no-subgraph} & 50.00 & 50.00 & 50.00 & 75.00 & 16.67 & 27.27 & 100.00 & 65.79 & 79.37 \\
\texttt{(baseline) random subgraph selection} & 60.71 & 100.00 & 75.56 & 52.94 & 50.00 & 51.43 & 100.00 & 76.32 & 86.57 \\
\texttt{(baseline) similarity-based ranker} & 62.96 & 100.00 & 77.27 & 50.00 & 61.11 & \textbf{55.00} & 100.00 & 78.95 & 88.24 \\
\texttt{(baseline) GPT-based ranker} & 62.96 & 100.00 & 77.27 & 50.00 & 55.56 & 52.63 & 100.00 & 81.58 & \textbf{89.86} \\
\rowcolor[HTML]{ECF4FF}\texttt{(ours) Subgraph Ranker (rmse-cross encoder)} & 62.96 & 100.00 & 77.27 & 47.37 & 50.00 & 48.65 & 100.00 & 86.84 & \underline{\textbf{92.96}} \\
\rowcolor[HTML]{ECF4FF}\texttt{(ours) Subgraph Ranker (ranknet-cross encoder)} & 65.38 & 100.00 & \underline{\textbf{79.07}} & 57.89 & 61.11 & \underline{\textbf{59.46}} & 100.00 & 81.58 & \textbf{89.86} \\
\rowcolor[HTML]{ECF4FF}\texttt{(ours) Subgraph Ranker (listnet-cross encoder)} & 62.96 & 100.00 & \textbf{78.27} & 52.63 & 55.56 & 54.05 & 100.00 & 81.58 & \textbf{89.86} \\
\rowcolor[HTML]{ECF4FF}\texttt{(ours) Subgraph Ranker (XGBoost)} & 60.71 & 100.00 & 75.56 & 56.25 & 50.00 & 52.94 & 100.00 & 73.68 & 84.85 \\
\bottomrule
\end{tabular}
\vspace{-3mm}
 \caption{Precision (P), Recall (R), and F1 scores on biomedical datasets. Values marked with \underline{underline} and \textbf{bold} represent the top and second-best F1 scores per dataset under different LLMs. The subgraphs were extracted from the \code{Hetionet}~\cite{hetionet10.7554/eLife.26726} KG.}
 \medskip
\label{tab:result}
\end{table*}

\subsection{\textbf{Method Comparison}}
\label{sec:modelcomparison} 
We compare the following methods: Models 1--4 serve as baselines, including a model without any knowledge from KGs and models enhanced with subgraphs ranked using baseline methods. Models 5--8, labeled ``\texttt{Subgraph Ranker}'', represent our proposed approach, where the model is provided with subgraphs ranked by variations of our proposed \srank.

\begin{enumerate}[leftmargin=0.5cm]
    \item [(1)] \textbf{\texttt{no-subgraph}}: only the variable pair, without any subgraphs.
    \item [(2)]\textbf{\texttt{random subgraph selection}}: providing the models with randomly selected subgraphs.
    \item [(3)] \textbf{\texttt{similarity-based ranker}}: text similarity-based method~\cite{surgekang2023knowledgegraphaugmentedlanguagemodels,TODshen-etal-2023-retrieval,KAPINGbaek-etal-2023-knowledge-augmented} adapted to subgraph ranking, detailed in Appendix \ref{app:simranker}. 
    \item [(4)] \textbf{\texttt{GPT-based ranker}}: instructing LLMs (\code{GPT}) to directly rank subgraphs, adapted from~\cite{sun-etal-2023-chatgpt} and detailed in Appendix \ref{app:gptranker}.  
    \item [(5)] \textbf{\texttt{Subgraph Ranker (rmse-cross encoder)}}: Our proposed \srank based on cross encoder model with \textit{pointwise} objective function \code{RMSE}, detailed in Appendix \ref{app:crossencoder}.
    \item [(6)] \textbf{\texttt{Subgraph Ranker (ranknet-cross encoder)}}: Our proposed \srank based on cross encoder model with \textit{pairwise} objective function \code{RankNet}, detailed in Appendix \ref{app:crossencoder}.
    \item [(7)] \textbf{\texttt{Subgraph Ranker (listnet-cross encoder)}}: Our proposed \srank based on cross encoder model with \textit{listwise} objective function \code{ListNet}, detailed in Appendix \ref{app:crossencoder}.
    \item [(8)] \textbf{\texttt{Subgraph Ranker (XGBoost)}}: Our proposed \srank based on gradient boosting \code{XGBoost}\cite{xgb10.1145/2939672.2939785}, see Appendix \ref{app:xgboost}. %
\end{enumerate}

\section{Results and Discussion}
\label{sec:resultdis}
Table~\ref{tab:result} \&~\ref{tab:result_sem} summarize the evaluation results for the biomedical and open-domain datasets. To summarize, methods incorporating subgraphs from KGs consistently outperformed the no-subgraph baseline across all experiments and datasets, with improvements of up to 44.4 points in F1 scores, evaluated using three different LLMs. For subgraph ranking, our proposed \srank emerged as the top-performing model in nearly all experiments, except in one case where the GPT-based ranker outperforms it. We provide detailed analysis and discussions of the results in the following.

\begin{table*}[tb]
\small
\centering
\vspace{-2mm}
\begin{tabular}{l|rrr|rrr|rrr}
\toprule
& \multicolumn{1}{c}{P} & \multicolumn{1}{c}{R} & \multicolumn{1}{c|}{F1} & \multicolumn{1}{c}{P} & \multicolumn{1}{c}{R} & \multicolumn{1}{c|}{F1} & \multicolumn{1}{c}{P} & \multicolumn{1}{c}{R} & \multicolumn{1}{c}{F1} \\
 \midrule
& \multicolumn{3}{c|}{\textcolor{blue}{\texttt{mistral-7b-instruct}}} & \multicolumn{3}{c|}{\textcolor{blue}{\texttt{llama-3-8b-instruct}}} & \multicolumn{3}{c}{{\textcolor{blue}{\texttt{gemma-7b-it}}}}  \\
\midrule
\texttt{(baseline) no-subgraph} & 60.00 & 25.00 & 35.29 & 71.43 & 41.67 & 52.63 & 50.00 & 91.67 & 64.71 \\
\texttt{(baseline) random subgraph selection} & 50.00 & 33.33 & 40.00 & 55.00 & 91.67 & 68.75 & 55.00 & 91.67 & 68.75 \\
\texttt{(baseline) similarity-based ranker} & 71.43 & 41.67 & \textbf{52.63} & 57.14 & 100.00 & \textbf{72.73} & 55.00 & 91.67 & 68.75 \\
\texttt{(baseline) GPT-based ranker} & 71.43 & 41.67 & \textbf{52.63} & 50.00 & 91.67 & 64.71 & 57.89 & 91.67 & \textbf{70.97} \\

\rowcolor[HTML]{ECF4FF}\texttt{(ours) Subgraph Ranker (rmse-cross encoder)} & 66.67 & 50.00 & \underline{\textbf{57.14}} & 60.00 & 100.00 & \underline{\textbf{75.00}} & 57.89 & 91.67 & \textbf{70.97} \\
\rowcolor[HTML]{ECF4FF}\texttt{(ours) Subgraph Ranker (ranknet-cross encoder)} & 66.67 & 33.33 & 44.44 & 57.89 & 91.67 & 70.97 & 57.89 & 91.67 & \textbf{70.97} \\
\rowcolor[HTML]{ECF4FF}\texttt{(ours) Subgraph Ranker (listnet-cross encoder)} & 57.14 & 33.33 & 42.11 & 57.89 & 91.67 & 70.97 & 57.89 & 91.67 & \textbf{70.97} \\
\rowcolor[HTML]{ECF4FF}\texttt{(ours) Subgraph Ranker (XGBoost)} & 55.56 & 41.67 & 47.62 & 57.14 & 100.00 & \textbf{72.73} & 61.11 & 91.67 & \underline{\textbf{73.33}} \\
\bottomrule
\end{tabular}
\vspace{-2mm}
 \caption{Precision (P), Recall (R), and F1 scores on open-domain dataset \code{SEMEVAL}\cite{SEMhendrickx-etal-2010-semeval}, with subgraphs from \code{Wikidata}~\cite{wikidata10.1145/2629489}.}
\label{tab:result_sem}
\end{table*}

\begin{table*}[tb]
\small
\centering
\vspace{-1mm}
\begin{tabular}{l|rrrrr}
\toprule
& \multicolumn{1}{c}{P $\uparrow$} & \multicolumn{1}{c}{R $\uparrow$} & \multicolumn{1}{c}{F1 $\uparrow$} & \multicolumn{1}{c}{HD $\downarrow$} & \multicolumn{1}{c}{NHD $\downarrow$} \\
\midrule 
\texttt{(statistical-based) PC Algorithm~\cite{PC10.7551/mitpress/1754.001.0001}} & 38.00 & 52.25 & 44.00 & 24 & 0.198\\
\texttt{(statistical-based) Exact Search (A*)~\cite{exact10.5555/2591248.2591250}} & 18.00 & 31.84 & 23.00 & 31 & 0.256\\
\texttt{(statistical-based) DirectLingam~\cite{lingam10.5555/1953048.2021040}} & 28.00 & 50.40 & 36.00 & 29 & 0.240\\

\texttt{(LLM-based-baseline) no-subgraph} & 100.00 & 15.79 & 27.27 & 16 & 0.132 \\
\rowcolor[HTML]{ECF4FF}\texttt{(ours, LLM+KG) Subgraph Ranker (rmse-cross encoder)} & 100.00 & 42.11 & 59.20 & 12 & 0.100 \\
\rowcolor[HTML]{ECF4FF}\texttt{(ours, LLM+KG) Subgraph Ranker (ranknet-cross encoder)} & 100.00 & 31.58 & 48.00 & 13 & 0.107 \\
\rowcolor[HTML]{ECF4FF}\texttt{(ours, LLM+KG) Subgraph Ranker (listnet-cross encoder)} & 100.00 & 63.16 & \underline{\textbf{77.42}} & \underline{\textbf{7}} & \underline{\textbf{0.057}} \\
\rowcolor[HTML]{ECF4FF}\texttt{(ours, LLM+KG) Subgraph Ranker (XGBoost)} & 100.00 & 42.11 & \textbf{59.26} & \textbf{11} & \textbf{0.091} \\
\bottomrule
\end{tabular}
\vspace{-2mm}
 \caption{Additional experiments on \texttt{SACHS}~\cite{sachsdoi:10.1126/science.1105809} observational data. }
\label{tab:result_sachs}
\end{table*}

\begin{discussion}{\textbf{Subgraph vs. No-Subgraph: Does the Subgraph Help LLMs Infer Causality?}} 
Methods incorporating subgraphs consistently outperform no-subgraph baselines across experiments and datasets. This evaluation, conducted with three different LLMs, shows significant performance improvements in knowledge-based causal discovery when enhanced with knowledge from KGs. The most notable result was observed with the \code{llama-3-8b} model on the \code{COMAGC} dataset, where our proposed approach achieved an F1 score of $77.78$ compared to $33.33$ for the no-subgraph baseline--a difference of 44.4 points in F1 scores. In addition, even randomly selected subgraphs, i.e., the \code{random subgraph selection} method, achieved better scores than the no-subgraph approach, with an average improvement of 12.5 points in F1 scores. This shows that metadata alone in prompts is not sufficient for accurate causal inference, and that integrating KGs significantly aids LLMs in inferring causality between variable pairs, leading to more accurate and reliable causal discovery.
\end{discussion}

\begin{discussion}{\textbf{Our \srank vs. LLM (GPT)-Based Ranker: How to Best Select Informative Subgraphs?}} In almost all experiments, at least one variation of our \srank is the best-performing model, except in one case 
where GPT-based ranker outperforms it ($83.87\:vs.\:87.50$ on \code{COMAGC} dataset with \code{mistral-7b}). This shows that our \srank excels in subgraph selection due to its specialized training on LLM-generated subgraph ranking data produced by \sre method. Additionally, incorporating \textit{meta} information from metapath-based subgraphs (e.g., node types) into the specialized model further optimized it for the ranking task. 
On the other hand, GPT-based ranker adapts the \textit{permutation generation}~\cite{sun-etal-2023-chatgpt} method for subgraph ranking by instructing GPT to rank subgraphs %
for a given 
variable pair.
While easy to implement, it is less effective than our approach, as LLMs like GPT are not optimized for specific ranking tasks.

To conclude, our proposed \srank offers an effective solution for selecting informative subgraphs from KGs to guide LLMs in knowledge-based causal discovery: it combines the strengths of LLMs in assessing subgraph \textit{informativeness} with a specialized trained model to produce an optimal ranking.
\end{discussion}

\begin{discussion}{\textbf{Our \srank vs. Similarity-Based Ranker.}}
In tasks such as QA and dialogue generation, similarity-based methods (e.g., cosine similarity of embeddings) 
is often employed to rank relevant information from KGs to ground the LLMs’ responses~\cite{surgekang2023knowledgegraphaugmentedlanguagemodels,TODshen-etal-2023-retrieval,KAPINGbaek-etal-2023-knowledge-augmented}. 
However, our experimental results show that this method (\code{similarity-\allowbreak based ranker}) is less effective than our approach for ranking subgraphs in KGs for knowledge-based causal discovery, 
with an average difference of 7.74 points in F1. This discrepancy likely arises because inferring causal relationship involves understanding deeper, often non-linear relationships between nodes, which similarity-based methods may fail to capture adequately. 
\end{discussion}

\begin{discussion}{\textbf{Cross Encoder vs. Gradient Boosting: Neural or Classical Subgraph Ranker?}} Our cross encoder-based \srank uses neural networks to rank the subgraphs, relying on contextual embeddings generated by the encoder model. In contrast, our gradient boosting-based \srank ranks subgraphs using decision trees that learn from manually defined features, such as \textit{n-grams}. Overall, while cross encoder-based subgraphs rankers generally deliver superior performance (9 out of 12 cases), the gradient boosting-based ranker performs relatively well in our task, emerging as the top model in 2 out of 12 cases (\code{SEMEVAL} with \code{gemma-7b} and \code{ADE} with \code{llama-3-8b}). According to~\cite{gavito2023gradientboostedbasedstructuredunstructured}, gradient-boosting excels on structured data, and although the subgraphs are represented as token sequences, they retain their structured nature from KGs, ensuring effective ranking. Meanwhile, within the variation of the LTR-based cross encoders, the pairwise approach (\code{ranknet}) consistently performs best, followed by pointwise (\code{rmse}) and listwise (\code{listnet}). The pairwise approach allows the model to learn subtle differences between subgraphs, leading to better ranking prediction.

\end{discussion}

\begin{discussion}{\textbf{Wikidata vs. Hetionet.}} 
For the biomedical datasets, using subgraphs from the \code{Hetionet} KG improved F1 scores by up to 44.4 points over the no-subgraph baseline. For open-domain dataset using \code{Wikidata}, the improvement was as high as 22.37 points, indicating its flexibility regarding the choice of KGs. The smaller improvement with \code{Wikidata} suggests that the LLMs already possess partial knowledge of certain relationships, %
likely learned from similar sources such as Wikipedia. However, our approach enhances this by guiding the model to focus on relevant causal paths, resulting in improved performance.
\end{discussion}

\subsection{\textbf{Comparison to Statistical-Based Causal Discovery}}
\label{sec:compstat}
Conventional causal discovery uses statistical methods on observational data, while our approach leverages LLMs and knowledge graphs to infer causal relationships from metadata.
Thus, to comprehensively evaluate our approach, we conducted additional experiments comparing it to conventional statistical-based methods. Specifically, we evaluated on the causal benchmark \texttt{SACHS}~\cite{sachsdoi:10.1126/science.1105809} protein causality dataset, comparing our method against conventional statistical-based causal discovery methods: \code{PC Algorithm}~\cite{PC10.7551/mitpress/1754.001.0001}, \code{Exact Search/A*}~\cite{exact10.5555/2591248.2591250}, and \code{DirectLingam}~\cite{lingam10.5555/1953048.2021040}.
The dataset was selected for its suitability in this evaluation scenario, as it contains real-world observational data needed for statistical-based analysis and a ground truth causal graph created by human experts, allowing empirical evaluation. The results are summarized in Table~\ref{tab:result_sachs}. Some of the statistical-based scores are obtained from~\cite{takayama2024integratinglargelanguagemodels}.

To sum up, our approach consistently outperforms statistical-based methods, achieving up to a 33.42-point improvement in F1 scores (\(77.42\) \textit{vs.} \(44.00\)) and a 17-point reduction in Hamming Distance (HD; the number of mismatches in the inferred adjacency matrix) (\(7\) \textit{vs.} \(24\)).
This demonstrates that LLMs, when integrated with structured information from KGs, effectively leverage semantic and contextual information embedded within the KGs, which allows them to infer causal relationship more accurately than traditional statistical methods. Unlike purely statistical approaches that rely heavily on observational datasets, our method does not require such data, making it more versatile and adaptable in real-world applications where observational data may be scarce or unavailable.

\section{Conclusion}
\label{sec:conclusion}
In this paper, we introduced a novel approach integrating KGs with LLMs to enhance knowledge-based causal discovery. Specifically, we utilized LLMs to identify informative subgraphs within KGs relevant to causal relationships and explored Learning-to-Rank-based specialized subgraph ranker models to further refine the subgraph selection. Our experiments revealed that our approach outperforms most baselines and performs consistently well across various LLMs and KGs on zero-shot knowledge-based causal discovery. 

Our research has focused on pairwise causal relationship; in future work, we aim to explore complex scenarios involving multiple interconnected variables, i.e., \textit{full} causal graph discovery. We also plan to co-optimize the integration of LLMs and KGs, while incorporating causality-related evaluation metrics to gain deeper insights into the underlying causal structures.

%% file: 3-appendix.tex
\appendix
\renewcommand\thefigure{\Alph{section}\arabic{figure}} 
\setcounter{figure}{0} 
\renewcommand\thetable{\Alph{section}\arabic{table}} 
\setcounter{table}{0} 

\paragraph{\noindent The following sections constitute the appendix.}

\section{Dataset Details}
\label{app:dataset}
The datasets are summarized in Table~\ref{tab:datastat}. Each instance in the dataset includes textual context where a variable pair co-occurs in a text, and is annotated by human experts to determine if there is a causal relation between the variable pair, as shown in the example.
\begin{itemize}[leftmargin=0.2cm]
    \item [] Example: \textit{\textbf{FGF6} contributes to the growth of \textbf{prostate cancer}}
    \item [] Ground truth: \code{(FGF6, causal, prostate cancer)}
\end{itemize}

\begin{table}[ht]
    \centering
    \caption{Dataset sizes and types.}
    \label{tab:datastat}
    \begin{tabular}{llr}
    \toprule
      \textbf{dataset} & \textbf{domain/type} & \textbf{total instances} \\
    \midrule
    \texttt{GENEC}~\cite{SusantiKGSP} & biomedical/gene-gene & 789\\
    \texttt{ADE}~\cite{ADEGURULINGAPPA2012885}  &  biomedical/drug-side effect &6,821 \\
    \texttt{COMAGC}~\cite{COMAGCLee2013com} &  biomedical/gene-disease &820 \\
    \texttt{SEMEVAL}~\cite{SEMhendrickx-etal-2010-semeval}  &  open-domain causality & 10,717 \\
    \bottomrule
    \end{tabular}
    \vspace{-2mm}
\end{table}

In the experiments, we filter the datasets (available in Github) to include only the instances that contain subgraph in the KGs. In addition, for the experiment in Table~\ref{tab:result_sachs}, we used \texttt{SACHS}~\cite{sachsdoi:10.1126/science.1105809} protein causality dataset, which contains 11 continuous variables.

\section{KG Querying \& Subgraph Extraction}
\label{app:kg}
We accessed the \textbf{Hetionet} KG through its official public Neo4j API\footnote{\url{bolt://neo4j.het.io}}, using \code{neo4j.v1} Python library. Neo4j is a third-party graph database for querying and visualizing knowledge graphs. We sampled the subgraphs from KGs based on \textbf{hop distance}; we queried up to 4 hops for extracting the subgraphs from Hetionet, using the built-in functions from Neo4J Cypher \code{allShortestPaths}. 

For \textbf{Wikidata}, we accessed the knowledge graph through its official public SPARQL endpoint\footnote{\url{https://query.wikidata.org/sparql}} using the \code{SPARQLWrapper} Python library. We employed the official Wikidata API (e.g., \code{wbsearchentities} and \code{wbgetentities} functions) to extract Wikidata IDs for all variable pairs. To extract subgraphs from \texttt{Wikidata}, we used the \textbf{shortest-path} method by~\cite{API10.1145/3488560.3498488}. We queried Wikidata to retrieve the subgraphs using the \texttt{CLOCQ}\footnote{\url{https://clocq.mpi-inf.mpg.de/documentation}} API, setting the maximum hops to 2. 

Since we will further rank these subgraphs, we performed \textbf{\textit{sparse retrieval} extraction} methods for each pair in the initial step before ranking, i.e., using \textbf{keyword-based/pattern-matching} querying. For instance, in Hetionet, to identify \textit{drugs} that target \textit{genes} related to specific \textit{diseases}, we query the KG using pattern such as: \code{(Drug)-[:TARGETS]->} \code{(Gene)-[:ASSOCIATED\_WITH]->(Disease)}. For Wikidata, we performed random sparse retrieval in this initial step.

\section{\sre: Details}
\label{app:sre}
We conducted \sre with \code{mistral-7b} from \code{huggingface}~\cite{huggingface-wolf-etal-2020-transformers} models library. To calculate subgraph relevance, we derive the log probability of tokens using beam transition scores, which are based on the log probabilities of tokens conditioned on the log softmax of previously generated tokens within the same beam. Example of the prompt and the generated subgraph ranking dataset are provided in the following:
\begin{tcolorbox}[boxsep=2pt,left=2pt,right=2pt,top=2pt,bottom=2pt]
\code{Given the following information, classify the relation between the drug and side effect. If there is a cause-effect relationship, state causal; otherwise, state non-causal. \\

[Pair]:
dihydrotachysterol and hypercalcemia

[Textual context]:
Severe hypercalcemia in a patient treated for hypoparathyroidism with dihydrotachysterol.

[Relation Paths]: FGF6 - tendon - SDRDL - FGFR2 - prostate cancer

[Relation]: }
\end{tcolorbox}

\begin{tcolorbox}[boxsep=2pt,left=2pt,right=2pt,top=2pt,bottom=2pt]
\code{{"qid": "1414", "e1": "carbamazepine", "e2": "systemic lupus erythematosus", "groundtruth": "1", "metapaths": \\

[{"pathid": 1, "relscore": 1.41565, "probscore": -0.7063895, "relevant": "1", "stops": "Carbamazepine - Conjunctivitis - Dasatinib - Systemic lupus erythematosus rash", "reltypes": "CAUSESCcSE - CAUSESCcSE - CAUSESCcSE", "nodelabels": "Compound - SideEffect - Compound - SideEffect"}, \\

{"pathid": 2, "relscore": -1.53165, "probscore": -0.6528874, "relevant": "0", "stops": "Carbamazepine - Renal failure - Dasatinib - Systemic lupus erythematosus rash", "reltypes": "CAUSESCcSE - CAUSESCcSE - CAUSESCcSE", "nodelabels": "Compound - SideEffect - Compound - SideEffect"}]}
}
\end{tcolorbox}

\section{\srank: Details}
\label{app:srank}

\subsection{Cross Encoder-Based Subgraph Ranker}
\label{app:crossencoder}
We experiment with the following models for training the cross encoders-based subgraph rankers: (1) \code{roberta-base}, (2) \code{deberta-base}, (3) \code{biomed\_roberta\_base}. We use the models obtained from the \code{huggingface}~\cite{huggingface-wolf-etal-2020-transformers} models library, and use their default hyperparamater settings to train the cross encoder. We use the implementation of the loss functions from \code{allRank}~\cite{AllRankPobrotyn2020AwareLT}\footnote{https://github.com/allegro/allRank/tree/master/allrank/models/losses}. All experiments were implemented in Python with \code{Pytorch} and \code{Transfomer} library.

We provided up to 10 subgraphs per pair for subgraph ranker training,  excluding pairs without any metapath-based subgraphs in our selected KGs. We concatenated the pair and its subgraph as the input to train the ranker, as shown in Eq.~\ref{eq:concat}. We added the \code{CLS} to mark the beginning of the sequence and \code{SEP} to separate the variable pair and the subgraph. When the \textit{node types} are not available, we used the \textit{relation labels} as the \textit{meta} information. The following shows the example of an input, where we included the \textit{node/variable name} and \textit{node types} as features:
\begin{tcolorbox}[boxsep=2pt,left=2pt,right=2pt,top=2pt,bottom=2pt]
\code{ CLS FGF6 prostate cancer SEP \gene{Gene} FGF6 - \anatomy{anatomy} tendon - \gene{gene} SDRDL - \gene{gene} FGFR2 - \disease{disease} prostate cancer }
\end{tcolorbox} 

Since the performance of the cross-encoder with different encoder models did not vary significantly, we selected \code{roberta-base} to train the cross encoder-based subgraph rankers. We provide the evaluation results of the cross encoder-based subgraph rankers (with \code{roberta-base} as the encoder) in Table~\ref{tab:app_srank}. Note that we used the ranking estimation of the subgraphs produced by \sre as ground truth to calculate the scores. 

\begin{table}[h]
\small
\centering
\begin{tabular}{l|l|r|r|r}
\toprule
 data & metric & \texttt{rmse} & \texttt{ranknet} & \texttt{listnet} \\
 \midrule
\multirow{3}{*}{\texttt{COMAGC}} & NDCG@5 & 56.60 & 57.12 & 55.38 \\
 & Recall@5 & 71.43 & 71.43 & 71.43 \\
 \midrule
\multirow{3}{*}{\texttt{GENEC}} & NDCG@5 & 56.79 & 55.89 & 55.52 \\
 & Recall@5 & 71.77 & 71.77 & 71.77 \\
 \midrule
\multirow{3}{*}{\texttt{ADE}} & NDCG@5 & 57.61 & 55.54 & 56.64 \\
 & Recall@5 & 72.18 & 72.18 & 72.18 \\
 \midrule
\multirow{3}{*}{\texttt{SEMEVAL}} & NDCG@5 & 91.92 & 89.42 & 88.99 \\
 & Recall@5 & 100.00 & 100.00 & 100.00 \\
\bottomrule
\end{tabular}
\caption{Subgraph Ranker experiment results.}
\label{tab:app_srank}
\end{table}

\subsection{XGBoost-Based Subgraph Ranker}
\label{app:xgboost}
We train our \code{xgboost}-based subgraph ranker as follows:
\begin{procedure}{\textbf{N-gram language model.}}
    This step trains an n-gram language model to generate embeddings as features for \code{xgboost}-based subgraph ranker training. During preprocessing, each variable pair in the dataset is concatenated with its corresponding subgraphs, combining all their connecting subgraphs into a single text sequence, as shown below. 
    \begin{tcolorbox}[boxsep=2pt,left=2pt,right=2pt,top=2pt,bottom=2pt]
    \code{
    carbamazepine - systemic lupus erythematosus: carbamazepine - renal failure - dasatinib - systemic lupus erythematosus}
    \end{tcolorbox}
    All concatenated sequences are then combined into a single corpus of text, which is used to generate n-grams. Using these n-grams, a simple neural network-based language model is built and trained to predict the next token in a sequence, learning the context and relationships between tokens. We further extract the embeddings from the n-gram model as features to train an \code{xgboost}-based subgraph ranker. We used $n$=2-gram and an embedding dimension of 128 to train the language model throughout the experiments. 
\end{procedure}

\begin{procedure}{\textbf{XGBoost-based subgraph ranker training.}} We implemented a gradient boosting-based subgraph ranker using the \code{XGBoost} Python package.\footnote{\url{https://github.com/dmlc/xgboost/}} For each sample in the dataset, we concatenated the variable pair with its corresponding subgraphs and extracted their embeddings using the trained n-gram language model as features to train the ranker models. Table~\ref{tab:xgboost} summarizes the hyperparameter values for \code{XGBoost}-based subgraph ranker training. 

\begin{table}[ht]
\small
\centering
  \begin{tabular}{lc}
    \toprule
    \textit{hyperparameter} & value \\
    \midrule    
    \code{tree method} & \code{hist} \\
    \code{lambdarank num pair per sample} & 8 \\
    \code{lambdarank pair method} & \code{topk} \\
    \code{objective} & \code{rank:ndcg} \\
    \bottomrule
  \end{tabular}
  \caption{\code{XGBoost}-based ranker hyperparameter values.}
  \label{tab:xgboost}
\end{table}
\end{procedure}

\section{\kbcd: Details}
\label{app:kbcd}
The previous module (\srank) produces ranked subgraphs; this step selects the top-$k$ most \textit{informative} subgraphs and incorporates them into the zero-shot prompt (details in \S\ref{sec:kbcd}). In our experiments, we found that including \textbf{up to one} ($k$=1) subgraph in the prompt yielded the best average results across all LLMs. We provide examples of the prompt in our Github.

\section{Baseline Implementation Details}
\label{app:baseline}
\subsection{Similarity-Based Subgraph Ranker}
\label{app:simranker}
In the \texttt{similarity-based ranker} method, we obtained embeddings for each pair and subgraph using the \code{Sentence Transformer}~\cite{reimers-2019-sentence-bert} pretrained model \code{all-MiniLM-L6-v2}, and then calculated the cosine similarity scores between 
them. 
The subgraphs were subsequently ranked based on these scores.

\subsection{GPT-Based Subgraph Ranker}
\label{app:gptranker}
RankGPT~\cite{sun-etal-2023-chatgpt} proposes a \textit{permutation generation} approach for passage re-ranking. This method involves passing a list of documents directly to an LLM and instructing it to rank them based on their relevance to a search query. We adapted RankGPT's original implementation\footnote{\url{https://github.com/sunnweiwei/RankGPT}} to rank subgraphs using a GPT model, providing up to 10 subgraphs per pair. We used the OpenAI API with the \code{gpt-3.5-turbo-instruct} engine for the GPT-based ranker. Generally, we assume only query access to the LLMs. Note that RankGPT also introduced a specialized passage ranker using permutation generation outputs; in our adaptation, we directly used the ranking output from permutation generation to rank the subgraphs. The following example shows the prompt for subgraph ranking:

\begin{tcolorbox}[boxsep=2pt,left=2pt,right=2pt,top=2pt,bottom=2pt]
\code{
role: system, content: You are an intelligent assistant that rank paths based on relevancy to entity pair. \\
role: user, content: I will provide you with 3 paths, each indicated by number identifier []. Rank the paths based on their usefulness in inferring causal relationships between the pair: \gene{(TRPV6, prostate cancer)}. \\
role: assistant, content: Okay, please provide the paths. \\
role: user, content: \anatomy{[1] TRPV6 - vagina - SQRDL - Lenalidomide - Prostate cancer metastatic} \\
role: user, content: \anatomy{[2] TRPV6 - TRAF1 - prostate cancer} \\
role: user, content: \anatomy{[3] TRPV6 - seminal vesicle - prostate cancer} \\
role: user, content: pair: \gene{(TRPV6, prostate cancer)}. Rank the 3 paths above based on their usefulness in inferring causal relationships between entity pair...
}
\end{tcolorbox}

\section{Computational Complexity and Hardware}
\label{app:comp-hw}
Our method comprises 3 modules with the following complexities:
\begin{itemize}[leftmargin=0.5cm]
    \item \textbf{Module 1 (\sre):} Complexity: $O(n \cdot p)$, where $n$ = subgraphs, $p$ = LLM parameters. Each subgraph requires a separate inference pass, ~5–15s on 7–8B LLMs.
    
    \item \textbf{Module 2 (\srank):} Subgraph ranking via (i) Transformer cross-encoder: $O(n \cdot m \cdot L^2)$, or (ii) XGBoost: $O(k \cdot d \cdot n \log(n))$, with $m$ = layers, $L$ = input length, $k$ = trees, $d$ = depth.
    
    \item \textbf{Module 3:} Zero-shot causal discovery. Complexity: $O(p)$. Each variable pair requires ~5–15s on 7–8B LLMs.
\end{itemize}

\noindent \textbf{Hardware specification:}
\begin{itemize}[leftmargin=0.5cm]
    \item 2× Tesla P100-PCIE (16GB each), Driver: 440.33.01, CUDA: 10.2
\end{itemize}

%% file: main.bbl

\begin{thebibliography}{65}


\ifx \showCODEN    \undefined \def \showCODEN     #1{\unskip}     \fi
\ifx \showISBNx    \undefined \def \showISBNx     #1{\unskip}     \fi
\ifx \showISBNxiii \undefined \def \showISBNxiii  #1{\unskip}     \fi
\ifx \showISSN     \undefined \def \showISSN      #1{\unskip}     \fi
\ifx \showLCCN     \undefined \def \showLCCN      #1{\unskip}     \fi
\ifx \shownote     \undefined \def \shownote      #1{#1}          \fi
\ifx \showarticletitle \undefined \def \showarticletitle #1{#1}   \fi
\ifx \showURL      \undefined \def \showURL       {\relax}        \fi
\providecommand\bibfield[2]{#2}
\providecommand\bibinfo[2]{#2}
\providecommand\natexlab[1]{#1}
\providecommand\showeprint[2][]{arXiv:#2}

\bibitem[Abdulaal et~al\mbox{.}(2024)]%
        {2024iclrcausal}
\bibfield{author}{\bibinfo{person}{Ahmed Abdulaal}, \bibinfo{person}{adamos hadjivasiliou}, \bibinfo{person}{Nina Montana-Brown}, \bibinfo{person}{Tiantian He}, \bibinfo{person}{Ayodeji Ijishakin}, \bibinfo{person}{Ivana Drobnjak}, \bibinfo{person}{Daniel~C. Castro}, {and} \bibinfo{person}{Daniel~C. Alexander}.} \bibinfo{year}{2024}\natexlab{}.
\newblock \showarticletitle{Causal Modelling Agents: Causal Graph Discovery through Synergising Metadata- and Data-driven Reasoning}. In \bibinfo{booktitle}{\emph{ICLR 2024 poster}}. \bibinfo{publisher}{International Conference on Learning Representations}, \bibinfo{address}{Kigali, Rwanda}, \bibinfo{pages}{123--130}.
\newblock


\bibitem[Baek et~al\mbox{.}(2023a)]%
        {KAPINGbaek-etal-2023-knowledge-augmented}
\bibfield{author}{\bibinfo{person}{Jinheon Baek}, \bibinfo{person}{Alham~Fikri Aji}, {and} \bibinfo{person}{Amir Saffari}.} \bibinfo{year}{2023}\natexlab{a}.
\newblock \showarticletitle{Knowledge-Augmented Language Model Prompting for Zero-Shot Knowledge Graph Question Answering}. In \bibinfo{booktitle}{\emph{Proceedings of MATCHING 2023}}, \bibfield{editor}{\bibinfo{person}{Estevam Hruschka}, \bibinfo{person}{Tom Mitchell}, \bibinfo{person}{Sajjadur Rahman}, \bibinfo{person}{Dunja Mladeni{\'c}}, {and} \bibinfo{person}{Marko Grobelnik}} (Eds.). \bibinfo{publisher}{Association for Computational Linguistics}, \bibinfo{address}{Toronto, ON, Canada}, \bibinfo{pages}{70--98}.
\newblock
\href{https://doi.org/10.18653/v1/2023.matching-1.7}{doi:\nolinkurl{10.18653/v1/2023.matching-1.7}}


\bibitem[Baek et~al\mbox{.}(2023b)]%
        {KALMVbaek-etal-2023-knowledge-augmented-language}
\bibfield{author}{\bibinfo{person}{Jinheon Baek}, \bibinfo{person}{Soyeong Jeong}, \bibinfo{person}{Minki Kang}, \bibinfo{person}{Jong Park}, {and} \bibinfo{person}{Sung Hwang}.} \bibinfo{year}{2023}\natexlab{b}.
\newblock \showarticletitle{Knowledge-Augmented Language Model Verification}. In \bibinfo{booktitle}{\emph{Proceedings of the 2023 Conference on Empirical Methods in Natural Language Processing}}, \bibfield{editor}{\bibinfo{person}{Houda Bouamor}, \bibinfo{person}{Juan Pino}, {and} \bibinfo{person}{Kalika Bali}} (Eds.). \bibinfo{publisher}{Association for Computational Linguistics}, \bibinfo{address}{Singapore}, \bibinfo{pages}{1720--1736}.
\newblock
\href{https://doi.org/10.18653/v1/2023.emnlp-main.107}{doi:\nolinkurl{10.18653/v1/2023.emnlp-main.107}}


\bibitem[Ban et~al\mbox{.}(2023)]%
        {ban2023querytoolscausalarchitects}
\bibfield{author}{\bibinfo{person}{Taiyu Ban}, \bibinfo{person}{Lyvzhou Chen}, \bibinfo{person}{Xiangyu Wang}, {and} \bibinfo{person}{Huanhuan Chen}.} \bibinfo{year}{2023}\natexlab{}.
\newblock \bibinfo{title}{From Query Tools to Causal Architects: Harnessing Large Language Models for Advanced Causal Discovery from Data}.
\newblock
\showeprint[arxiv]{2306.16902}~[cs.AI]
\urldef\tempurl%
\url{https://arxiv.org/abs/2306.16902}
\showURL{%
\tempurl}


\bibitem[Burges(2010)]%
        {Burges2010FromRT}
\bibfield{author}{\bibinfo{person}{Christopher J.~C. Burges}.} \bibinfo{year}{2010}\natexlab{}.
\newblock \bibinfo{booktitle}{\emph{From RankNet to LambdaRank to LambdaMART: An Overview}}.
\newblock \bibinfo{type}{{T}echnical {R}eport} MSR-TR-2010-82. \bibinfo{institution}{Microsoft Research}.
\newblock


\bibitem[Cao et~al\mbox{.}(2007)]%
        {Listnetcao2007learning}
\bibfield{author}{\bibinfo{person}{Zhe Cao}, \bibinfo{person}{Tao Qin}, \bibinfo{person}{Tie-Yan Liu}, \bibinfo{person}{Ming-Feng Tsai}, {and} \bibinfo{person}{Hang Li}.} \bibinfo{year}{2007}\natexlab{}.
\newblock \bibinfo{booktitle}{\emph{Learning to Rank: From Pairwise Approach to Listwise Approach}}.
\newblock \bibinfo{type}{{T}echnical {R}eport} MSR-TR-2007-40. \bibinfo{institution}{Microsoft Research}. \bibinfo{pages}{9} pages.
\newblock


\bibitem[Chang et~al\mbox{.}(2024)]%
        {Chang_2024path}
\bibfield{author}{\bibinfo{person}{Heng Chang}, \bibinfo{person}{Jiangnan Ye}, \bibinfo{person}{Alejo Lopez-Avila}, \bibinfo{person}{Jinhua Du}, {and} \bibinfo{person}{Jia Li}.} \bibinfo{year}{2024}\natexlab{}.
\newblock \showarticletitle{Path-based Explanation for Knowledge Graph Completion}. In \bibinfo{booktitle}{\emph{Proceedings of the 30th ACM SIGKDD Conference on Knowledge Discovery and Data Mining}} \emph{(\bibinfo{series}{KDD ’24})}. \bibinfo{publisher}{ACM}, \bibinfo{pages}{231–242}.
\newblock
\href{https://doi.org/10.1145/3637528.3671683}{doi:\nolinkurl{10.1145/3637528.3671683}}


\bibitem[Chatwal et~al\mbox{.}(2025)]%
        {chatwal-etal-2025-enhancing}
\bibfield{author}{\bibinfo{person}{Pulkit Chatwal}, \bibinfo{person}{Amit Agarwal}, {and} \bibinfo{person}{Ankush Mittal}.} \bibinfo{year}{2025}\natexlab{}.
\newblock \showarticletitle{Enhancing Causal Relationship Detection Using Prompt Engineering and Large Language Models}. In \bibinfo{booktitle}{\emph{Proceedings of the Joint Workshop of the 9th Financial Technology and Natural Language Processing (FinNLP), the 6th Financial Narrative Processing (FNP), and the 1st Workshop on Large Language Models for Finance and Legal (LLMFinLegal)}}, \bibfield{editor}{\bibinfo{person}{Chung-Chi Chen}, \bibinfo{person}{Antonio Moreno-Sandoval}, \bibinfo{person}{Jimin Huang}, \bibinfo{person}{Qianqian Xie}, \bibinfo{person}{Sophia Ananiadou}, {and} \bibinfo{person}{Hsin-Hsi Chen}} (Eds.). \bibinfo{publisher}{Association for Computational Linguistics}, \bibinfo{address}{Abu Dhabi, UAE}, \bibinfo{pages}{248--252}.
\newblock
\urldef\tempurl%
\url{https://aclanthology.org/2025.finnlp-1.26/}
\showURL{%
\tempurl}


\bibitem[Chen and Guestrin(2016)]%
        {xgb10.1145/2939672.2939785}
\bibfield{author}{\bibinfo{person}{Tianqi Chen} {and} \bibinfo{person}{Carlos Guestrin}.} \bibinfo{year}{2016}\natexlab{}.
\newblock \showarticletitle{XGBoost: A Scalable Tree Boosting System}. In \bibinfo{booktitle}{\emph{Proceedings of the 22nd ACM SIGKDD International Conference on Knowledge Discovery and Data Mining}} (San Francisco, California, USA) \emph{(\bibinfo{series}{KDD '16})}. \bibinfo{publisher}{Association for Computing Machinery}, \bibinfo{address}{New York, NY, USA}, \bibinfo{pages}{785–794}.
\newblock
\showISBNx{9781450342322}
\href{https://doi.org/10.1145/2939672.2939785}{doi:\nolinkurl{10.1145/2939672.2939785}}


\bibitem[Christmann et~al\mbox{.}(2022)]%
        {API10.1145/3488560.3498488}
\bibfield{author}{\bibinfo{person}{Philipp Christmann}, \bibinfo{person}{Rishiraj Saha~Roy}, {and} \bibinfo{person}{Gerhard Weikum}.} \bibinfo{year}{2022}\natexlab{}.
\newblock \showarticletitle{Beyond NED: Fast and Effective Search Space Reduction for Complex Question Answering over Knowledge Bases}. In \bibinfo{booktitle}{\emph{Proceedings of the Fifteenth ACM International Conference on Web Search and Data Mining}} (Virtual Event, AZ, USA) \emph{(\bibinfo{series}{WSDM '22})}. \bibinfo{publisher}{Association for Computing Machinery}, \bibinfo{address}{New York, NY, USA}, \bibinfo{pages}{172–180}.
\newblock
\showISBNx{9781450391320}
\href{https://doi.org/10.1145/3488560.3498488}{doi:\nolinkurl{10.1145/3488560.3498488}}


\bibitem[Devlin et~al\mbox{.}(2018)]%
        {BERTDBLP:journals/corr/abs-1810-04805}
\bibfield{author}{\bibinfo{person}{Jacob Devlin}, \bibinfo{person}{Ming{-}Wei Chang}, \bibinfo{person}{Kenton Lee}, {and} \bibinfo{person}{Kristina Toutanova}.} \bibinfo{year}{2018}\natexlab{}.
\newblock \showarticletitle{{BERT:} Pre-training of Deep Bidirectional Transformers for Language Understanding}.
\newblock \bibinfo{journal}{\emph{CoRR}}  \bibinfo{volume}{abs/1810.04805} (\bibinfo{year}{2018}).
\newblock
\showeprint[arXiv]{1810.04805}
\urldef\tempurl%
\url{http://arxiv.org/abs/1810.04805}
\showURL{%
\tempurl}


\bibitem[Feder et~al\mbox{.}(2022)]%
        {10.1162/tacl_a_00511}
\bibfield{author}{\bibinfo{person}{Amir Feder}, \bibinfo{person}{Katherine~A. Keith}, \bibinfo{person}{Emaad Manzoor}, \bibinfo{person}{Reid Pryzant}, \bibinfo{person}{Dhanya Sridhar}, \bibinfo{person}{Zach Wood-Doughty}, \bibinfo{person}{Jacob Eisenstein}, \bibinfo{person}{Justin Grimmer}, \bibinfo{person}{Roi Reichart}, \bibinfo{person}{Margaret~E. Roberts}, \bibinfo{person}{Brandon~M. Stewart}, \bibinfo{person}{Victor Veitch}, {and} \bibinfo{person}{Diyi Yang}.} \bibinfo{year}{2022}\natexlab{}.
\newblock \showarticletitle{{Causal Inference in Natural Language Processing: Estimation, Prediction, Interpretation and Beyond}}.
\newblock \bibinfo{journal}{\emph{Transactions of the Association for Computational Linguistics}}  \bibinfo{volume}{10} (\bibinfo{date}{10} \bibinfo{year}{2022}), \bibinfo{pages}{1138--1158}.
\newblock
\showISSN{2307-387X}
\urldef\tempurl%
\url{https://doi.org/10.1162/tacl\_a\_00511}
\showURL{%
\tempurl}


\bibitem[Gao et~al\mbox{.}(2023)]%
        {gao-etal-2023-chatgpt}
\bibfield{author}{\bibinfo{person}{Jinglong Gao}, \bibinfo{person}{Xiao Ding}, \bibinfo{person}{Bing Qin}, {and} \bibinfo{person}{Ting Liu}.} \bibinfo{year}{2023}\natexlab{}.
\newblock \showarticletitle{Is {C}hat{GPT} a Good Causal Reasoner? A Comprehensive Evaluation}. In \bibinfo{booktitle}{\emph{Findings of the Association for Computational Linguistics: EMNLP 2023}}, \bibfield{editor}{\bibinfo{person}{Houda Bouamor}, \bibinfo{person}{Juan Pino}, {and} \bibinfo{person}{Kalika Bali}} (Eds.). \bibinfo{publisher}{Association for Computational Linguistics}, \bibinfo{address}{Singapore}, \bibinfo{pages}{11111--11126}.
\newblock
\href{https://doi.org/10.18653/v1/2023.findings-emnlp.743}{doi:\nolinkurl{10.18653/v1/2023.findings-emnlp.743}}


\bibitem[Gao et~al\mbox{.}(2021)]%
        {bertrank10.1007/978-3-030-72240-1_26}
\bibfield{author}{\bibinfo{person}{Luyu Gao}, \bibinfo{person}{Zhuyun Dai}, {and} \bibinfo{person}{Jamie Callan}.} \bibinfo{year}{2021}\natexlab{}.
\newblock \showarticletitle{Rethink Training of BERT Rerankers in Multi-stage Retrieval Pipeline}. In \bibinfo{booktitle}{\emph{Advances in Information Retrieval: 43rd European Conference on IR Research, ECIR 2021, Virtual Event, March 28 – April 1, 2021, Proceedings, Part II}}. \bibinfo{publisher}{Springer-Verlag}, \bibinfo{address}{Berlin, Heidelberg}, \bibinfo{pages}{280–286}.
\newblock
\showISBNx{978-3-030-72239-5}
\href{https://doi.org/10.1007/978-3-030-72240-1_26}{doi:\nolinkurl{10.1007/978-3-030-72240-1_26}}


\bibitem[Gavito et~al\mbox{.}(2023)]%
        {gavito2023gradientboostedbasedstructuredunstructured}
\bibfield{author}{\bibinfo{person}{Andrea~Treviño Gavito}, \bibinfo{person}{Diego Klabjan}, {and} \bibinfo{person}{Jean Utke}.} \bibinfo{year}{2023}\natexlab{}.
\newblock \bibinfo{title}{Gradient-Boosted Based Structured and Unstructured Learning}.
\newblock
\showeprint[arxiv]{2302.14299}~[cs.LG]
\urldef\tempurl%
\url{https://arxiv.org/abs/2302.14299}
\showURL{%
\tempurl}


\bibitem[Gurulingappa et~al\mbox{.}(2012)]%
        {ADEGURULINGAPPA2012885}
\bibfield{author}{\bibinfo{person}{Harsha Gurulingappa}, \bibinfo{person}{Abdul~Mateen Rajput}, \bibinfo{person}{Angus Roberts}, \bibinfo{person}{Juliane Fluck}, \bibinfo{person}{Martin Hofmann-Apitius}, {and} \bibinfo{person}{Luca Toldo}.} \bibinfo{year}{2012}\natexlab{}.
\newblock \showarticletitle{Development of a benchmark corpus to support the automatic extraction of drug-related adverse effects from medical case reports}.
\newblock \bibinfo{journal}{\emph{Journal of Biomedical Informatics}} \bibinfo{volume}{45}, \bibinfo{number}{5} (\bibinfo{year}{2012}), \bibinfo{pages}{885 -- 892}.
\newblock
\showISSN{1532-0464}
\href{https://doi.org/10.1016/j.jbi.2012.04.008}{doi:\nolinkurl{10.1016/j.jbi.2012.04.008}}
\newblock
\shownote{Text Mining and Natural Language Processing in Pharmacogenomics}.


\bibitem[Hendrickx et~al\mbox{.}(2010)]%
        {SEMhendrickx-etal-2010-semeval}
\bibfield{author}{\bibinfo{person}{Iris Hendrickx}, \bibinfo{person}{Su~Nam Kim}, \bibinfo{person}{Zornitsa Kozareva}, \bibinfo{person}{Preslav Nakov}, \bibinfo{person}{Diarmuid {\'O}~S{\'e}aghdha}, \bibinfo{person}{Sebastian Pad{\'o}}, \bibinfo{person}{Marco Pennacchiotti}, \bibinfo{person}{Lorenza Romano}, {and} \bibinfo{person}{Stan Szpakowicz}.} \bibinfo{year}{2010}\natexlab{}.
\newblock \showarticletitle{{S}em{E}val-2010 Task 8: Multi-Way Classification of Semantic Relations between Pairs of Nominals}. In \bibinfo{booktitle}{\emph{Proceedings of the 5th International Workshop on Semantic Evaluation}}. \bibinfo{publisher}{Association for Computational Linguistics}, \bibinfo{address}{Uppsala, Sweden}, \bibinfo{pages}{33--38}.
\newblock
\urldef\tempurl%
\url{https://aclanthology.org/S10-1006}
\showURL{%
\tempurl}


\bibitem[Himmelstein et~al\mbox{.}(2017)]%
        {hetionet10.7554/eLife.26726}
\bibfield{author}{\bibinfo{person}{Daniel~Scott Himmelstein}, \bibinfo{person}{Antoine Lizee}, \bibinfo{person}{Christine Hessler}, \bibinfo{person}{Leo Brueggeman}, \bibinfo{person}{Sabrina~L Chen}, \bibinfo{person}{Dexter Hadley}, \bibinfo{person}{Ari Green}, \bibinfo{person}{Pouya Khankhanian}, {and} \bibinfo{person}{Sergio~E Baranzini}.} \bibinfo{year}{2017}\natexlab{}.
\newblock \showarticletitle{Systematic integration of biomedical knowledge prioritizes drugs for repurposing}.
\newblock \bibinfo{journal}{\emph{eLife}}  \bibinfo{volume}{6} (\bibinfo{date}{sep} \bibinfo{year}{2017}), \bibinfo{pages}{e26726}.
\newblock
\showISSN{2050-084X}
\href{https://doi.org/10.7554/eLife.26726}{doi:\nolinkurl{10.7554/eLife.26726}}


\bibitem[Hong et~al\mbox{.}(2023)]%
        {hong2023faithfulquestionansweringmontecarlo}
\bibfield{author}{\bibinfo{person}{Ruixin Hong}, \bibinfo{person}{Hongming Zhang}, \bibinfo{person}{Hong Zhao}, \bibinfo{person}{Dong Yu}, {and} \bibinfo{person}{Changshui Zhang}.} \bibinfo{year}{2023}\natexlab{}.
\newblock \bibinfo{title}{Faithful Question Answering with Monte-Carlo Planning}.
\newblock
\showeprint[arxiv]{2305.02556}~[cs.CL]
\urldef\tempurl%
\url{https://arxiv.org/abs/2305.02556}
\showURL{%
\tempurl}


\bibitem[Jiang et~al\mbox{.}(2023)]%
        {jiang2023mistral7b}
\bibfield{author}{\bibinfo{person}{Albert~Q. Jiang}, \bibinfo{person}{Alexandre Sablayrolles}, \bibinfo{person}{Arthur Mensch}, \bibinfo{person}{Chris Bamford}, \bibinfo{person}{Devendra~Singh Chaplot}, \bibinfo{person}{Diego de~las Casas}, \bibinfo{person}{Florian Bressand}, \bibinfo{person}{Gianna Lengyel}, \bibinfo{person}{Guillaume Lample}, \bibinfo{person}{Lucile Saulnier}, \bibinfo{person}{Lélio~Renard Lavaud}, \bibinfo{person}{Marie-Anne Lachaux}, \bibinfo{person}{Pierre Stock}, \bibinfo{person}{Teven~Le Scao}, \bibinfo{person}{Thibaut Lavril}, \bibinfo{person}{Thomas Wang}, \bibinfo{person}{Timothée Lacroix}, {and} \bibinfo{person}{William~El Sayed}.} \bibinfo{year}{2023}\natexlab{}.
\newblock \bibinfo{title}{Mistral 7B}.
\newblock
\showeprint[arxiv]{2310.06825}~[cs.CL]
\urldef\tempurl%
\url{https://arxiv.org/abs/2310.06825}
\showURL{%
\tempurl}


\bibitem[Jiang et~al\mbox{.}(2024)]%
        {hykgehypothesisknowledgegraph}
\bibfield{author}{\bibinfo{person}{Xinke Jiang}, \bibinfo{person}{Ruizhe Zhang}, \bibinfo{person}{Yongxin Xu}, \bibinfo{person}{Rihong Qiu}, \bibinfo{person}{Yue Fang}, \bibinfo{person}{Zhiyuan Wang}, \bibinfo{person}{Jinyi Tang}, \bibinfo{person}{Hongxin Ding}, \bibinfo{person}{Xu Chu}, \bibinfo{person}{Junfeng Zhao}, {and} \bibinfo{person}{Yasha Wang}.} \bibinfo{year}{2024}\natexlab{}.
\newblock \bibinfo{title}{HyKGE: A Hypothesis Knowledge Graph Enhanced Framework for Accurate and Reliable Medical LLMs Responses}.
\newblock
\showeprint[arxiv]{2312.15883}
\urldef\tempurl%
\url{https://arxiv.org/abs/2312.15883}
\showURL{%
\tempurl}


\bibitem[Jiralerspong et~al\mbox{.}(2024)]%
        {jiralerspong2024efficientcausalgraphdiscovery}
\bibfield{author}{\bibinfo{person}{Thomas Jiralerspong}, \bibinfo{person}{Xiaoyin Chen}, \bibinfo{person}{Yash More}, \bibinfo{person}{Vedant Shah}, {and} \bibinfo{person}{Yoshua Bengio}.} \bibinfo{year}{2024}\natexlab{}.
\newblock \bibinfo{title}{Efficient Causal Graph Discovery Using Large Language Models}.
\newblock
\showeprint[arxiv]{2402.01207}~[cs.LG]
\urldef\tempurl%
\url{https://arxiv.org/abs/2402.01207}
\showURL{%
\tempurl}


\bibitem[Kang et~al\mbox{.}(2023)]%
        {surgekang2023knowledgegraphaugmentedlanguagemodels}
\bibfield{author}{\bibinfo{person}{Minki Kang}, \bibinfo{person}{Jin~Myung Kwak}, \bibinfo{person}{Jinheon Baek}, {and} \bibinfo{person}{Sung~Ju Hwang}.} \bibinfo{year}{2023}\natexlab{}.
\newblock \bibinfo{title}{Knowledge Graph-Augmented Language Models for Knowledge-Grounded Dialogue Generation}.
\newblock
\showeprint[arxiv]{2305.18846}~[cs.CL]
\urldef\tempurl%
\url{https://arxiv.org/abs/2305.18846}
\showURL{%
\tempurl}


\bibitem[Khetan et~al\mbox{.}(2022)]%
        {khetan-etal-2022-mimicause}
\bibfield{author}{\bibinfo{person}{Vivek Khetan}, \bibinfo{person}{Md~Imbesat Rizvi}, \bibinfo{person}{Jessica Huber}, \bibinfo{person}{Paige Bartusiak}, \bibinfo{person}{Bogdan Sacaleanu}, {and} \bibinfo{person}{Andrew Fano}.} \bibinfo{year}{2022}\natexlab{}.
\newblock \showarticletitle{{MIMIC}ause: {R}epresentation and automatic extraction of causal relation types from clinical notes}. In \bibinfo{booktitle}{\emph{Findings of the Association for Computational Linguistics: ACL 2022}}. \bibinfo{publisher}{Association for Computational Linguistics}, \bibinfo{address}{Dublin, Ireland}, \bibinfo{pages}{764--773}.
\newblock
\href{https://doi.org/10.18653/v1/2022.findings-acl.63}{doi:\nolinkurl{10.18653/v1/2022.findings-acl.63}}


\bibitem[Kıcıman et~al\mbox{.}(2023)]%
        {kic2023causal}
\bibfield{author}{\bibinfo{person}{Emre Kıcıman}, \bibinfo{person}{Robert Ness}, \bibinfo{person}{Amit Sharma}, {and} \bibinfo{person}{Chenhao Tan}.} \bibinfo{year}{2023}\natexlab{}.
\newblock \bibinfo{title}{Causal Reasoning and Large Language Models: Opening a New Frontier for Causality}.
\newblock


\bibitem[Lee et~al\mbox{.}(2013)]%
        {COMAGCLee2013com}
\bibfield{author}{\bibinfo{person}{Hee-Jin Lee}, \bibinfo{person}{Sang-Hyung Shim}, \bibinfo{person}{Mi-Ryoung Song}, \bibinfo{person}{Hyunju Lee}, {and} \bibinfo{person}{Jong~C. Park}.} \bibinfo{year}{2013}\natexlab{}.
\newblock \showarticletitle{CoMAGC: a corpus with multi-faceted annotations of gene-cancer relations}.
\newblock \bibinfo{journal}{\emph{BMC Bioinformatics}} \bibinfo{volume}{14}, \bibinfo{number}{1} (\bibinfo{date}{Nov} \bibinfo{year}{2013}), \bibinfo{pages}{323}.
\newblock
\showISSN{1471-2105}
\urldef\tempurl%
\url{https://doi.org/10.1186/1471-2105-14-323}
\showURL{%
\tempurl}


\bibitem[Lewis et~al\mbox{.}(2020)]%
        {rag10.5555/3495724.3496517}
\bibfield{author}{\bibinfo{person}{Patrick Lewis}, \bibinfo{person}{Ethan Perez}, \bibinfo{person}{Aleksandra Piktus}, \bibinfo{person}{Fabio Petroni}, \bibinfo{person}{Vladimir Karpukhin}, \bibinfo{person}{Naman Goyal}, \bibinfo{person}{Heinrich K\"{u}ttler}, \bibinfo{person}{Mike Lewis}, \bibinfo{person}{Wen-tau Yih}, \bibinfo{person}{Tim Rockt\"{a}schel}, \bibinfo{person}{Sebastian Riedel}, {and} \bibinfo{person}{Douwe Kiela}.} \bibinfo{year}{2020}\natexlab{}.
\newblock \showarticletitle{Retrieval-augmented generation for knowledge-intensive NLP tasks}. In \bibinfo{booktitle}{\emph{Proceedings of the 34th International Conference on Neural Information Processing Systems}} (Vancouver, BC, Canada) \emph{(\bibinfo{series}{NIPS '20})}. \bibinfo{publisher}{Curran Associates Inc.}, \bibinfo{address}{Red Hook, NY, USA}, Article \bibinfo{articleno}{793}, \bibinfo{numpages}{16}~pages.
\newblock
\showISBNx{9781713829546}


\bibitem[Li et~al\mbox{.}(2024)]%
        {7bli2024common7blanguagemodels}
\bibfield{author}{\bibinfo{person}{Chen Li}, \bibinfo{person}{Weiqi Wang}, \bibinfo{person}{Jingcheng Hu}, \bibinfo{person}{Yixuan Wei}, \bibinfo{person}{Nanning Zheng}, \bibinfo{person}{Han Hu}, \bibinfo{person}{Zheng Zhang}, {and} \bibinfo{person}{Houwen Peng}.} \bibinfo{year}{2024}\natexlab{}.
\newblock \bibinfo{title}{Common 7B Language Models Already Possess Strong Math Capabilities}.
\newblock
\showeprint[arxiv]{2403.04706}
\urldef\tempurl%
\url{https://arxiv.org/abs/2403.04706}
\showURL{%
\tempurl}


\bibitem[Liu(2009)]%
        {IR10.1561/1500000016}
\bibfield{author}{\bibinfo{person}{Tie-Yan Liu}.} \bibinfo{year}{2009}\natexlab{}.
\newblock \showarticletitle{Learning to Rank for Information Retrieval}.
\newblock \bibinfo{journal}{\emph{Found. Trends Inf. Retr.}} \bibinfo{volume}{3}, \bibinfo{number}{3} (\bibinfo{date}{mar} \bibinfo{year}{2009}), \bibinfo{pages}{225–331}.
\newblock
\showISSN{1554-0669}
\href{https://doi.org/10.1561/1500000016}{doi:\nolinkurl{10.1561/1500000016}}


\bibitem[Luo et~al\mbox{.}(2024)]%
        {luo2024rog}
\bibfield{author}{\bibinfo{person}{Linhao Luo}, \bibinfo{person}{Yuan-Fang Li}, \bibinfo{person}{Gholamreza Haffari}, {and} \bibinfo{person}{Shirui Pan}.} \bibinfo{year}{2024}\natexlab{}.
\newblock \showarticletitle{Reasoning on Graphs: Faithful and Interpretable Large Language Model Reasoning}. In \bibinfo{booktitle}{\emph{International Conference on Learning Representations (ICLR)}}.
\newblock


\bibitem[Meta({[n.\,d.]})]%
        {llama3}
\bibfield{author}{\bibinfo{person}{Meta}.} \bibinfo{year}{[n.\,d.]}\natexlab{}.
\newblock \bibinfo{title}{Introducing Meta Llama 3: The most capable openly available LLM to date}.
\newblock \bibinfo{howpublished}{\url{https://ai.meta.com/blog/meta-llama-3/}}.
\newblock
\newblock
\shownote{[Accessed 29-07-2024]}.


\bibitem[Nogueira and Cho(2019)]%
        {Nogueira2019PassageRW}
\bibfield{author}{\bibinfo{person}{Rodrigo Nogueira} {and} \bibinfo{person}{Kyunghyun Cho}.} \bibinfo{year}{2019}\natexlab{}.
\newblock \showarticletitle{Passage Re-ranking with BERT}.
\newblock \bibinfo{journal}{\emph{ArXiv}}  \bibinfo{volume}{abs/1901.04085} (\bibinfo{year}{2019}).
\newblock


\bibitem[Noori et~al\mbox{.}(2023)]%
        {meta10.1093/bioinformatics/btad297}
\bibfield{author}{\bibinfo{person}{Ayush Noori}, \bibinfo{person}{Michelle~M Li}, \bibinfo{person}{Amelia L~M Tan}, {and} \bibinfo{person}{Marinka Zitnik}.} \bibinfo{year}{2023}\natexlab{}.
\newblock \showarticletitle{{Metapaths: similarity search in heterogeneous knowledge graphs via meta-paths}}.
\newblock \bibinfo{journal}{\emph{Bioinformatics}} \bibinfo{volume}{39}, \bibinfo{number}{5} (\bibinfo{date}{05} \bibinfo{year}{2023}), \bibinfo{pages}{btad297}.
\newblock
\showISSN{1367-4811}
\urldef\tempurl%
\url{https://doi.org/10.1093/bioinformatics/btad297}
\showURL{%
\tempurl}


\bibitem[Pan et~al\mbox{.}(2024)]%
        {Pan_2024}
\bibfield{author}{\bibinfo{person}{Shirui Pan}, \bibinfo{person}{Linhao Luo}, \bibinfo{person}{Yufei Wang}, \bibinfo{person}{Chen Chen}, \bibinfo{person}{Jiapu Wang}, {and} \bibinfo{person}{Xindong Wu}.} \bibinfo{year}{2024}\natexlab{}.
\newblock \showarticletitle{Unifying Large Language Models and Knowledge Graphs: A Roadmap}.
\newblock \bibinfo{journal}{\emph{IEEE Transactions on Knowledge and Data Engineering}} \bibinfo{volume}{36}, \bibinfo{number}{7} (\bibinfo{date}{July} \bibinfo{year}{2024}), \bibinfo{pages}{3580–3599}.
\newblock
\showISSN{2326-3865}
\href{https://doi.org/10.1109/tkde.2024.3352100}{doi:\nolinkurl{10.1109/tkde.2024.3352100}}


\bibitem[Petroni et~al\mbox{.}(2020)]%
        {PromptPetroniLPRWM020}
\bibfield{author}{\bibinfo{person}{Fabio Petroni}, \bibinfo{person}{Patrick S.~H. Lewis}, \bibinfo{person}{Aleksandra Piktus}, \bibinfo{person}{Tim Rocktäschel}, \bibinfo{person}{Yuxiang Wu}, \bibinfo{person}{Alexander~H. Miller}, {and} \bibinfo{person}{Sebastian Riedel}.} \bibinfo{year}{2020}\natexlab{}.
\newblock \showarticletitle{How Context Affects Language Models' Factual Predictions.}. In \bibinfo{booktitle}{\emph{AKBC}}.
\newblock
\urldef\tempurl%
\url{http://dblp.uni-trier.de/db/conf/akbc/akbc2020.html#PetroniLPRWM020}
\showURL{%
\tempurl}


\bibitem[Pobrotyn et~al\mbox{.}(2020)]%
        {AllRankPobrotyn2020AwareLT}
\bibfield{author}{\bibinfo{person}{Przemyslaw Pobrotyn}, \bibinfo{person}{Tomasz Bartczak}, \bibinfo{person}{Mikolaj Synowiec}, \bibinfo{person}{Radoslaw Bialobrzeski}, {and} \bibinfo{person}{Jaroslaw Bojar}.} \bibinfo{year}{2020}\natexlab{}.
\newblock \showarticletitle{Context-Aware Learning to Rank with Self-Attention}. In \bibinfo{booktitle}{\emph{Proceedings of ACM SIGIR Workshop on eCommerce (SIGI ReCom}} (China (Virtual Event)).
\newblock


\bibitem[Qin et~al\mbox{.}(2021)]%
        {50030}
\bibfield{author}{\bibinfo{person}{Zhen Qin}, \bibinfo{person}{Le Yan}, \bibinfo{person}{Honglei Zhuang}, \bibinfo{person}{Yi Tay}, \bibinfo{person}{Rama~Kumar Pasumarthi}, \bibinfo{person}{Xuanhui Wang}, \bibinfo{person}{Mike Bendersky}, {and} \bibinfo{person}{Marc Najork}.} \bibinfo{year}{2021}\natexlab{}.
\newblock \showarticletitle{Are Neural Rankers still Outperformed by Gradient Boosted Decision Trees?}. In \bibinfo{booktitle}{\emph{International Conference on Learning Representations (ICLR)}}.
\newblock


\bibitem[Ranade and Joshi(2024)]%
        {fabula10.1145/3625007.3627505}
\bibfield{author}{\bibinfo{person}{Priyanka Ranade} {and} \bibinfo{person}{Anupam Joshi}.} \bibinfo{year}{2024}\natexlab{}.
\newblock \showarticletitle{FABULA: Intelligence Report Generation Using Retrieval-Augmented Narrative Construction}. In \bibinfo{booktitle}{\emph{Proceedings of the 2023 IEEE/ACM International Conference on Advances in Social Networks Analysis and Mining}} (Kusadasi, Turkiye) \emph{(\bibinfo{series}{ASONAM '23})}. \bibinfo{publisher}{Association for Computing Machinery}, \bibinfo{address}{New York, NY, USA}, \bibinfo{pages}{603–610}.
\newblock
\showISBNx{9798400704093}
\href{https://doi.org/10.1145/3625007.3627505}{doi:\nolinkurl{10.1145/3625007.3627505}}


\bibitem[Reimers and Gurevych(2019)]%
        {reimers-2019-sentence-bert}
\bibfield{author}{\bibinfo{person}{Nils Reimers} {and} \bibinfo{person}{Iryna Gurevych}.} \bibinfo{year}{2019}\natexlab{}.
\newblock \showarticletitle{Sentence-BERT: Sentence Embeddings using Siamese BERT-Networks}. In \bibinfo{booktitle}{\emph{Proceedings of the 2019 Conference on Empirical Methods in Natural Language Processing}}. \bibinfo{publisher}{Association for Computational Linguistics}.
\newblock
\urldef\tempurl%
\url{http://arxiv.org/abs/1908.10084}
\showURL{%
\tempurl}


\bibitem[Ren et~al\mbox{.}(2023)]%
        {Renmeta3}
\bibfield{author}{\bibinfo{person}{Zhong-Hao Ren}, \bibinfo{person}{Zhu-Hong You}, \bibinfo{person}{Quan Zou}, \bibinfo{person}{Chang-Qing Yu}, \bibinfo{person}{Yan-Fang Ma}, \bibinfo{person}{Yong-Jian Guan}, \bibinfo{person}{Hai-Ru You}, \bibinfo{person}{Xin-Fei Wang}, {and} \bibinfo{person}{Jie Pan}.} \bibinfo{year}{2023}\natexlab{}.
\newblock \showarticletitle{DeepMPF: deep learning framework for predicting drug--target interactions based on multi-modal representation with meta-path semantic analysis}.
\newblock \bibinfo{journal}{\emph{Journal of Translational Medicine}} \bibinfo{volume}{21}, \bibinfo{number}{1} (\bibinfo{date}{25 Jan} \bibinfo{year}{2023}), \bibinfo{pages}{48}.
\newblock
\showISSN{1479-5876}
\urldef\tempurl%
\url{https://doi.org/10.1186/s12967-023-03876-3}
\showURL{%
\tempurl}


\bibitem[Reynolds and McDonell(2021)]%
        {Prompt10.1145/3411763.3451760}
\bibfield{author}{\bibinfo{person}{Laria Reynolds} {and} \bibinfo{person}{Kyle McDonell}.} \bibinfo{year}{2021}\natexlab{}.
\newblock \showarticletitle{Prompt Programming for Large Language Models: Beyond the Few-Shot Paradigm}. In \bibinfo{booktitle}{\emph{Extended Abstracts of the 2021 CHI Conference on Human Factors in Computing Systems}} (Yokohama, Japan) \emph{(\bibinfo{series}{CHI EA '21})}. \bibinfo{publisher}{Association for Computing Machinery}, \bibinfo{address}{New York, NY, USA}, Article \bibinfo{articleno}{314}, \bibinfo{numpages}{7}~pages.
\newblock
\showISBNx{9781450380959}
\href{https://doi.org/10.1145/3411763.3451760}{doi:\nolinkurl{10.1145/3411763.3451760}}


\bibitem[Sachs et~al\mbox{.}(2005)]%
        {sachsdoi:10.1126/science.1105809}
\bibfield{author}{\bibinfo{person}{Karen Sachs}, \bibinfo{person}{Omar Perez}, \bibinfo{person}{Dana Pe'er}, \bibinfo{person}{Douglas~A. Lauffenburger}, {and} \bibinfo{person}{Garry~P. Nolan}.} \bibinfo{year}{2005}\natexlab{}.
\newblock \showarticletitle{Causal Protein-Signaling Networks Derived from Multiparameter Single-Cell Data}.
\newblock \bibinfo{journal}{\emph{Science}} \bibinfo{volume}{308}, \bibinfo{number}{5721} (\bibinfo{year}{2005}), \bibinfo{pages}{523--529}.
\newblock
\href{https://doi.org/10.1126/science.1105809}{doi:\nolinkurl{10.1126/science.1105809}}
\showeprint{https://www.science.org/doi/pdf/10.1126/science.1105809}


\bibitem[Shen et~al\mbox{.}(2023)]%
        {TODshen-etal-2023-retrieval}
\bibfield{author}{\bibinfo{person}{Weizhou Shen}, \bibinfo{person}{Yingqi Gao}, \bibinfo{person}{Canbin Huang}, \bibinfo{person}{Fanqi Wan}, \bibinfo{person}{Xiaojun Quan}, {and} \bibinfo{person}{Wei Bi}.} \bibinfo{year}{2023}\natexlab{}.
\newblock \showarticletitle{Retrieval-Generation Alignment for End-to-End Task-Oriented Dialogue System}. In \bibinfo{booktitle}{\emph{Proceedings of the 2023 Conference on Empirical Methods in Natural Language Processing}}, \bibfield{editor}{\bibinfo{person}{Houda Bouamor}, \bibinfo{person}{Juan Pino}, {and} \bibinfo{person}{Kalika Bali}} (Eds.). \bibinfo{publisher}{Association for Computational Linguistics}, \bibinfo{address}{Singapore}, \bibinfo{pages}{8261--8275}.
\newblock
\href{https://doi.org/10.18653/v1/2023.emnlp-main.514}{doi:\nolinkurl{10.18653/v1/2023.emnlp-main.514}}


\bibitem[Shimizu et~al\mbox{.}(2011)]%
        {lingam10.5555/1953048.2021040}
\bibfield{author}{\bibinfo{person}{Shohei Shimizu}, \bibinfo{person}{Takanori Inazumi}, \bibinfo{person}{Yasuhiro Sogawa}, \bibinfo{person}{Aapo Hyv\"{a}rinen}, \bibinfo{person}{Yoshinobu Kawahara}, \bibinfo{person}{Takashi Washio}, \bibinfo{person}{Patrik~O. Hoyer}, {and} \bibinfo{person}{Kenneth Bollen}.} \bibinfo{year}{2011}\natexlab{}.
\newblock \showarticletitle{DirectLiNGAM: A Direct Method for Learning a Linear Non-Gaussian Structural Equation Model}.
\newblock \bibinfo{journal}{\emph{J. Mach. Learn. Res.}} \bibinfo{volume}{12}, \bibinfo{number}{null} (\bibinfo{date}{July} \bibinfo{year}{2011}), \bibinfo{pages}{1225–1248}.
\newblock
\showISSN{1532-4435}


\bibitem[Spirtes et~al\mbox{.}(2000)]%
        {Spirtes2000}
\bibfield{author}{\bibinfo{person}{Peter Spirtes}, \bibinfo{person}{Clark Glymour}, {and} \bibinfo{person}{Scheines R}.} \bibinfo{year}{2000}\natexlab{}.
\newblock \showarticletitle{{Constructing bayesian networks models of gene expression networks from microarray data}}. In \bibinfo{booktitle}{\emph{{Proceedings of the Atlantic Symposium on Computational Biology}}} (North Carolina).
\newblock


\bibitem[Spirtes et~al\mbox{.}(2001a)]%
        {Spirtes2001}
\bibfield{author}{\bibinfo{person}{P. Spirtes}, \bibinfo{person}{C. Glymour}, {and} \bibinfo{person}{R. Scheines}.} \bibinfo{year}{2001}\natexlab{a}.
\newblock \bibinfo{booktitle}{\emph{Causation, Prediction, and Search} (\bibinfo{edition}{2nd} ed.)}.
\newblock \bibinfo{publisher}{MIT press}.
\newblock


\bibitem[Spirtes et~al\mbox{.}(2001b)]%
        {PC10.7551/mitpress/1754.001.0001}
\bibfield{author}{\bibinfo{person}{Peter Spirtes}, \bibinfo{person}{Clark Glymour}, {and} \bibinfo{person}{Richard Scheines}.} \bibinfo{year}{2001}\natexlab{b}.
\newblock \bibinfo{booktitle}{\emph{Causation, Prediction, and Search}}.
\newblock \bibinfo{publisher}{The MIT Press}.
\newblock
\showISBNx{9780262284158}
\href{https://doi.org/10.7551/mitpress/1754.001.0001}{doi:\nolinkurl{10.7551/mitpress/1754.001.0001}}


\bibitem[Sun et~al\mbox{.}(2023)]%
        {sun-etal-2023-chatgpt}
\bibfield{author}{\bibinfo{person}{Weiwei Sun}, \bibinfo{person}{Lingyong Yan}, \bibinfo{person}{Xinyu Ma}, \bibinfo{person}{Shuaiqiang Wang}, \bibinfo{person}{Pengjie Ren}, \bibinfo{person}{Zhumin Chen}, \bibinfo{person}{Dawei Yin}, {and} \bibinfo{person}{Zhaochun Ren}.} \bibinfo{year}{2023}\natexlab{}.
\newblock \showarticletitle{Is {C}hat{GPT} Good at Search? Investigating Large Language Models as Re-Ranking Agents}. In \bibinfo{booktitle}{\emph{Proceedings of the 2023 Conference on Empirical Methods in Natural Language Processing}}, \bibfield{editor}{\bibinfo{person}{Houda Bouamor}, \bibinfo{person}{Juan Pino}, {and} \bibinfo{person}{Kalika Bali}} (Eds.). \bibinfo{publisher}{Association for Computational Linguistics}, \bibinfo{address}{Singapore}, \bibinfo{pages}{14918--14937}.
\newblock
\href{https://doi.org/10.18653/v1/2023.emnlp-main.923}{doi:\nolinkurl{10.18653/v1/2023.emnlp-main.923}}


\bibitem[Sun et~al\mbox{.}(2011)]%
        {metapath10.14778/3402707.3402736}
\bibfield{author}{\bibinfo{person}{Yizhou Sun}, \bibinfo{person}{Jiawei Han}, \bibinfo{person}{Xifeng Yan}, \bibinfo{person}{Philip~S. Yu}, {and} \bibinfo{person}{Tianyi Wu}.} \bibinfo{year}{2011}\natexlab{}.
\newblock \showarticletitle{PathSim: meta path-based top-K similarity search in heterogeneous information networks}.
\newblock \bibinfo{journal}{\emph{Proc. VLDB Endow.}} \bibinfo{volume}{4}, \bibinfo{number}{11} (\bibinfo{date}{aug} \bibinfo{year}{2011}), \bibinfo{pages}{992–1003}.
\newblock
\showISSN{2150-8097}
\href{https://doi.org/10.14778/3402707.3402736}{doi:\nolinkurl{10.14778/3402707.3402736}}


\bibitem[Susanti and F{\"{a}}rber(2024)]%
        {SusantiKGSP}
\bibfield{author}{\bibinfo{person}{Yuni Susanti} {and} \bibinfo{person}{Michael F{\"{a}}rber}.} \bibinfo{year}{2024}\natexlab{}.
\newblock \showarticletitle{{Knowledge Graph Structure as Prompt: Improving Small Language Models Capabilities for Knowledge-based Causal Discovery}}. In \bibinfo{booktitle}{\emph{{Proceedings of the ISWC'24}}} (Baltimore, USA).
\newblock


\bibitem[Susanti and Uchino(2024)]%
        {SusantiCEG}
\bibfield{author}{\bibinfo{person}{Yuni Susanti} {and} \bibinfo{person}{Kanji Uchino}.} \bibinfo{year}{2024}\natexlab{}.
\newblock \showarticletitle{Causal-Evidence Graph for Causal Relation Classification}. In \bibinfo{booktitle}{\emph{Proceedings of the 39th ACM/SIGAPP Symposium on Applied Computing}} (Avila, Spain) \emph{(\bibinfo{series}{SAC '24})}. \bibinfo{publisher}{Association for Computing Machinery}, \bibinfo{address}{New York, NY, USA}, \bibinfo{pages}{714–722}.
\newblock
\showISBNx{9798400702433}
\href{https://doi.org/10.1145/3605098.3635894}{doi:\nolinkurl{10.1145/3605098.3635894}}


\bibitem[Takayama et~al\mbox{.}(2024)]%
        {takayama2024integratinglargelanguagemodels}
\bibfield{author}{\bibinfo{person}{Masayuki Takayama}, \bibinfo{person}{Tadahisa Okuda}, \bibinfo{person}{Thong Pham}, \bibinfo{person}{Tatsuyoshi Ikenoue}, \bibinfo{person}{Shingo Fukuma}, \bibinfo{person}{Shohei Shimizu}, {and} \bibinfo{person}{Akiyoshi Sannai}.} \bibinfo{year}{2024}\natexlab{}.
\newblock \bibinfo{title}{Integrating Large Language Models in Causal Discovery: A Statistical Causal Approach}.
\newblock
\showeprint[arxiv]{2402.01454}~[cs.LG]
\urldef\tempurl%
\url{https://arxiv.org/abs/2402.01454}
\showURL{%
\tempurl}


\bibitem[Team et~al\mbox{.}(2024)]%
        {gemmateam2024gemmaopenmodelsbased}
\bibfield{author}{\bibinfo{person}{Gemma Team}, \bibinfo{person}{Thomas Mesnard}, \bibinfo{person}{Cassidy Hardin}, \bibinfo{person}{Robert Dadashi}, \bibinfo{person}{Surya Bhupatiraju}, \bibinfo{person}{Shreya Pathak}, \bibinfo{person}{Laurent Sifre}, \bibinfo{person}{Morgane Rivière}, \bibinfo{person}{Mihir~Sanjay Kale}, \bibinfo{person}{Juliette Love}, \bibinfo{person}{Pouya Tafti}, \bibinfo{person}{Léonard Hussenot}, \bibinfo{person}{Pier~Giuseppe Sessa}, \bibinfo{person}{Aakanksha Chowdhery}, \bibinfo{person}{Adam Roberts}, \bibinfo{person}{Aditya Barua}, \bibinfo{person}{Alex Botev}, \bibinfo{person}{Alex Castro-Ros}, \bibinfo{person}{Ambrose Slone}, \bibinfo{person}{Amélie Héliou}, \bibinfo{person}{Andrea Tacchetti}, \bibinfo{person}{Anna Bulanova}, \bibinfo{person}{Antonia Paterson}, \bibinfo{person}{Beth Tsai}, \bibinfo{person}{Bobak Shahriari}, \bibinfo{person}{Charline~Le Lan}, \bibinfo{person}{Christopher~A. Choquette-Choo}, \bibinfo{person}{Clément Crepy}, \bibinfo{person}{Daniel Cer},
  \bibinfo{person}{Daphne Ippolito}, \bibinfo{person}{David Reid}, \bibinfo{person}{Elena Buchatskaya}, \bibinfo{person}{Eric Ni}, \bibinfo{person}{Eric Noland}, \bibinfo{person}{Geng Yan}, \bibinfo{person}{George Tucker}, \bibinfo{person}{George-Christian Muraru}, \bibinfo{person}{Grigory Rozhdestvenskiy}, \bibinfo{person}{Henryk Michalewski}, \bibinfo{person}{Ian Tenney}, \bibinfo{person}{Ivan Grishchenko}, \bibinfo{person}{Jacob Austin}, \bibinfo{person}{James Keeling}, \bibinfo{person}{Jane Labanowski}, \bibinfo{person}{Jean-Baptiste Lespiau}, \bibinfo{person}{Jeff Stanway}, \bibinfo{person}{Jenny Brennan}, \bibinfo{person}{Jeremy Chen}, \bibinfo{person}{Johan Ferret}, \bibinfo{person}{Justin Chiu}, \bibinfo{person}{Justin Mao-Jones}, \bibinfo{person}{Katherine Lee}, \bibinfo{person}{Kathy Yu}, \bibinfo{person}{Katie Millican}, \bibinfo{person}{Lars~Lowe Sjoesund}, \bibinfo{person}{Lisa Lee}, \bibinfo{person}{Lucas Dixon}, \bibinfo{person}{Machel Reid}, \bibinfo{person}{Maciej Mikuła},
  \bibinfo{person}{Mateo Wirth}, \bibinfo{person}{Michael Sharman}, \bibinfo{person}{Nikolai Chinaev}, \bibinfo{person}{Nithum Thain}, \bibinfo{person}{Olivier Bachem}, \bibinfo{person}{Oscar Chang}, \bibinfo{person}{Oscar Wahltinez}, \bibinfo{person}{Paige Bailey}, \bibinfo{person}{Paul Michel}, \bibinfo{person}{Petko Yotov}, \bibinfo{person}{Rahma Chaabouni}, \bibinfo{person}{Ramona Comanescu}, \bibinfo{person}{Reena Jana}, \bibinfo{person}{Rohan Anil}, \bibinfo{person}{Ross McIlroy}, \bibinfo{person}{Ruibo Liu}, \bibinfo{person}{Ryan Mullins}, \bibinfo{person}{Samuel~L Smith}, \bibinfo{person}{Sebastian Borgeaud}, \bibinfo{person}{Sertan Girgin}, \bibinfo{person}{Sholto Douglas}, \bibinfo{person}{Shree Pandya}, \bibinfo{person}{Siamak Shakeri}, \bibinfo{person}{Soham De}, \bibinfo{person}{Ted Klimenko}, \bibinfo{person}{Tom Hennigan}, \bibinfo{person}{Vlad Feinberg}, \bibinfo{person}{Wojciech Stokowiec}, \bibinfo{person}{Yu hui Chen}, \bibinfo{person}{Zafarali Ahmed}, \bibinfo{person}{Zhitao Gong},
  \bibinfo{person}{Tris Warkentin}, \bibinfo{person}{Ludovic Peran}, \bibinfo{person}{Minh Giang}, \bibinfo{person}{Clément Farabet}, \bibinfo{person}{Oriol Vinyals}, \bibinfo{person}{Jeff Dean}, \bibinfo{person}{Koray Kavukcuoglu}, \bibinfo{person}{Demis Hassabis}, \bibinfo{person}{Zoubin Ghahramani}, \bibinfo{person}{Douglas Eck}, \bibinfo{person}{Joelle Barral}, \bibinfo{person}{Fernando Pereira}, \bibinfo{person}{Eli Collins}, \bibinfo{person}{Armand Joulin}, \bibinfo{person}{Noah Fiedel}, \bibinfo{person}{Evan Senter}, \bibinfo{person}{Alek Andreev}, {and} \bibinfo{person}{Kathleen Kenealy}.} \bibinfo{year}{2024}\natexlab{}.
\newblock \bibinfo{title}{Gemma: Open Models Based on Gemini Research and Technology}.
\newblock
\showeprint[arxiv]{2403.08295}~[cs.CL]
\urldef\tempurl%
\url{https://arxiv.org/abs/2403.08295}
\showURL{%
\tempurl}


\bibitem[Thakkar and Manimaran(2023)]%
        {7b10434081}
\bibfield{author}{\bibinfo{person}{Hiren Thakkar} {and} \bibinfo{person}{A Manimaran}.} \bibinfo{year}{2023}\natexlab{}.
\newblock \showarticletitle{Comprehensive Examination of Instruction-Based Language Models: A Comparative Analysis of Mistral-7B and Llama-2-7B}. In \bibinfo{booktitle}{\emph{2023 International Conference on Emerging Research in Computational Science (ICERCS)}}. \bibinfo{pages}{1--6}.
\newblock
\href{https://doi.org/10.1109/ICERCS57948.2023.10434081}{doi:\nolinkurl{10.1109/ICERCS57948.2023.10434081}}


\bibitem[Tu et~al\mbox{.}(2023)]%
        {tu2023causaldiscovery}
\bibfield{author}{\bibinfo{person}{Ruibo Tu}, \bibinfo{person}{Chao Ma}, {and} \bibinfo{person}{Cheng Zhang}.} \bibinfo{year}{2023}\natexlab{}.
\newblock \bibinfo{title}{Causal-Discovery Performance of ChatGPT in the context of Neuropathic Pain Diagnosis}.
\newblock
\showeprint[arxiv]{2301.13819}~[cs.CL]


\bibitem[Vrande\v{c}i\'{c} and Kr\"{o}tzsch(2014)]%
        {wikidata10.1145/2629489}
\bibfield{author}{\bibinfo{person}{Denny Vrande\v{c}i\'{c}} {and} \bibinfo{person}{Markus Kr\"{o}tzsch}.} \bibinfo{year}{2014}\natexlab{}.
\newblock \showarticletitle{Wikidata: a free collaborative knowledgebase}.
\newblock \bibinfo{journal}{\emph{Commun. ACM}} \bibinfo{volume}{57}, \bibinfo{number}{10} (\bibinfo{date}{sep} \bibinfo{year}{2014}), \bibinfo{pages}{78–85}.
\newblock
\showISSN{0001-0782}
\urldef\tempurl%
\url{https://doi.org/10.1145/2629489}
\showURL{%
\tempurl}


\bibitem[Wang et~al\mbox{.}(2022)]%
        {wangmeta1}
\bibfield{author}{\bibinfo{person}{Hong Wang}, \bibinfo{person}{Xiaoqi Wang}, \bibinfo{person}{Wenjuan Liu}, \bibinfo{person}{Xiaolan Xie}, {and} \bibinfo{person}{Shaoliang Peng}.} \bibinfo{year}{2022}\natexlab{}.
\newblock \showarticletitle{deepDGA: Biomedical Heterogeneous Network-based Deep Learning Framework for Disease-Gene Association Predictions}. In \bibinfo{booktitle}{\emph{{IEEE} International Conference on Bioinformatics and Biomedicine, {BIBM} 2022, Las Vegas, NV, USA, December 6-8, 2022}}, \bibfield{editor}{\bibinfo{person}{Donald~A. Adjeroh}, \bibinfo{person}{Qi~Long}, \bibinfo{person}{Xinghua~Mindy Shi}, \bibinfo{person}{Fei Guo}, \bibinfo{person}{Xiaohua Hu}, \bibinfo{person}{Srinivas Aluru}, \bibinfo{person}{Giri Narasimhan}, \bibinfo{person}{Jianxin Wang}, \bibinfo{person}{Mingon Kang}, \bibinfo{person}{Ananda Mondal}, {and} \bibinfo{person}{Jin Liu}} (Eds.). \bibinfo{publisher}{{IEEE}}, \bibinfo{pages}{601--606}.
\newblock
\urldef\tempurl%
\url{https://doi.org/10.1109/BIBM55620.2022.9995651}
\showURL{%
\tempurl}


\bibitem[Wang et~al\mbox{.}(2023a)]%
        {wang2023knowledgedrivencotexploringfaithful}
\bibfield{author}{\bibinfo{person}{Keheng Wang}, \bibinfo{person}{Feiyu Duan}, \bibinfo{person}{Sirui Wang}, \bibinfo{person}{Peiguang Li}, \bibinfo{person}{Yunsen Xian}, \bibinfo{person}{Chuantao Yin}, \bibinfo{person}{Wenge Rong}, {and} \bibinfo{person}{Zhang Xiong}.} \bibinfo{year}{2023}\natexlab{a}.
\newblock \bibinfo{title}{Knowledge-Driven CoT: Exploring Faithful Reasoning in LLMs for Knowledge-intensive Question Answering}.
\newblock
\showeprint[arxiv]{2308.13259}~[cs.CL]
\urldef\tempurl%
\url{https://arxiv.org/abs/2308.13259}
\showURL{%
\tempurl}


\bibitem[Wang et~al\mbox{.}(2023b)]%
        {knowledgptenhancinglargelanguage}
\bibfield{author}{\bibinfo{person}{Xintao Wang}, \bibinfo{person}{Qianwen Yang}, \bibinfo{person}{Yongting Qiu}, \bibinfo{person}{Jiaqing Liang}, \bibinfo{person}{Qianyu He}, \bibinfo{person}{Zhouhong Gu}, \bibinfo{person}{Yanghua Xiao}, {and} \bibinfo{person}{Wei Wang}.} \bibinfo{year}{2023}\natexlab{b}.
\newblock \bibinfo{title}{KnowledGPT: Enhancing Large Language Models with Retrieval and Storage Access on Knowledge Bases}.
\newblock
\showeprint[arxiv]{2308.11761}~[cs.CL]
\urldef\tempurl%
\url{https://arxiv.org/abs/2308.11761}
\showURL{%
\tempurl}


\bibitem[Willig et~al\mbox{.}(2022)]%
        {willig2022foundation}
\bibfield{author}{\bibinfo{person}{Moritz Willig}, \bibinfo{person}{Matej Zečević}, \bibinfo{person}{Devendra~Singh Dhami}, {and} \bibinfo{person}{Kristian Kersting}.} \bibinfo{year}{2022}\natexlab{}.
\newblock \bibinfo{title}{Can Foundation Models Talk Causality?}
\newblock
\showeprint[arxiv]{2206.10591}~[cs.AI]


\bibitem[Wolf et~al\mbox{.}(2020)]%
        {huggingface-wolf-etal-2020-transformers}
\bibfield{author}{\bibinfo{person}{Thomas Wolf}, \bibinfo{person}{Lysandre Debut}, \bibinfo{person}{Victor Sanh}, \bibinfo{person}{Julien Chaumond}, \bibinfo{person}{Clement Delangue}, \bibinfo{person}{Anthony Moi}, \bibinfo{person}{Pierric Cistac}, \bibinfo{person}{Tim Rault}, \bibinfo{person}{Remi Louf}, \bibinfo{person}{Morgan Funtowicz}, \bibinfo{person}{Joe Davison}, \bibinfo{person}{Sam Shleifer}, \bibinfo{person}{Patrick von Platen}, \bibinfo{person}{Clara Ma}, \bibinfo{person}{Yacine Jernite}, \bibinfo{person}{Julien Plu}, \bibinfo{person}{Canwen Xu}, \bibinfo{person}{Teven Le~Scao}, \bibinfo{person}{Sylvain Gugger}, \bibinfo{person}{Mariama Drame}, \bibinfo{person}{Quentin Lhoest}, {and} \bibinfo{person}{Alexander Rush}.} \bibinfo{year}{2020}\natexlab{}.
\newblock \showarticletitle{Transformers: State-of-the-Art Natural Language Processing}. In \bibinfo{booktitle}{\emph{Proceedings of the 2020 Conference on Empirical Methods in Natural Language Processing: System Demonstrations}}, \bibfield{editor}{\bibinfo{person}{Qun Liu} {and} \bibinfo{person}{David Schlangen}} (Eds.). \bibinfo{publisher}{Association for Computational Linguistics}, \bibinfo{address}{Online}, \bibinfo{pages}{38--45}.
\newblock
\urldef\tempurl%
\url{https://aclanthology.org/2020.emnlp-demos.6}
\showURL{%
\tempurl}


\bibitem[Yao et~al\mbox{.}(2022)]%
        {yaometa2}
\bibfield{author}{\bibinfo{person}{Wenjie Yao}, \bibinfo{person}{Weizhong Zhao}, \bibinfo{person}{Xingpeng Jiang}, \bibinfo{person}{Xianjun Shen}, {and} \bibinfo{person}{Tingting He}.} \bibinfo{year}{2022}\natexlab{}.
\newblock \showarticletitle{{MPGNN-DSA:} {A} Meta-path-based Graph Neural Network for drug-side effect association prediction}. In \bibinfo{booktitle}{\emph{{IEEE} International Conference on Bioinformatics and Biomedicine, {BIBM} 2022, Las Vegas, NV, USA, December 6-8, 2022}}, \bibfield{editor}{\bibinfo{person}{Donald~A. Adjeroh}, \bibinfo{person}{Qi~Long}, \bibinfo{person}{Xinghua~Mindy Shi}, \bibinfo{person}{Fei Guo}, \bibinfo{person}{Xiaohua Hu}, \bibinfo{person}{Srinivas Aluru}, \bibinfo{person}{Giri Narasimhan}, \bibinfo{person}{Jianxin Wang}, \bibinfo{person}{Mingon Kang}, \bibinfo{person}{Ananda Mondal}, {and} \bibinfo{person}{Jin Liu}} (Eds.). \bibinfo{publisher}{{IEEE}}, \bibinfo{pages}{627--632}.
\newblock
\urldef\tempurl%
\url{https://doi.org/10.1109/BIBM55620.2022.9995486}
\showURL{%
\tempurl}


\bibitem[Yuan and Malone(2013)]%
        {exact10.5555/2591248.2591250}
\bibfield{author}{\bibinfo{person}{Changhe Yuan} {and} \bibinfo{person}{Brandon Malone}.} \bibinfo{year}{2013}\natexlab{}.
\newblock \showarticletitle{Learning optimal bayesian networks: a shortest path perspective}.
\newblock \bibinfo{journal}{\emph{J. Artif. Int. Res.}} \bibinfo{volume}{48}, \bibinfo{number}{1} (\bibinfo{date}{Oct.} \bibinfo{year}{2013}), \bibinfo{pages}{23–65}.
\newblock
\showISSN{1076-9757}


\bibitem[Zhang et~al\mbox{.}(2023a)]%
        {zhang2023understanding}
\bibfield{author}{\bibinfo{person}{Cheng Zhang}, \bibinfo{person}{Stefan Bauer}, \bibinfo{person}{Paul Bennett}, \bibinfo{person}{Jiangfeng Gao}, \bibinfo{person}{Wenbo Gong}, \bibinfo{person}{Agrin Hilmkil}, \bibinfo{person}{Joel Jennings}, \bibinfo{person}{Chao Ma}, \bibinfo{person}{Tom Minka}, \bibinfo{person}{Nick Pawlowski}, {and} \bibinfo{person}{James Vaughan}.} \bibinfo{year}{2023}\natexlab{a}.
\newblock \bibinfo{title}{Understanding Causality with Large Language Models: Feasibility and Opportunities}.
\newblock
\showeprint[arxiv]{2304.05524}~[cs.LG]


\bibitem[Zhang et~al\mbox{.}(2023b)]%
        {zhang2023pagelinkpathbasedgraphneural}
\bibfield{author}{\bibinfo{person}{Shichang Zhang}, \bibinfo{person}{Jiani Zhang}, \bibinfo{person}{Xiang Song}, \bibinfo{person}{Soji Adeshina}, \bibinfo{person}{Da Zheng}, \bibinfo{person}{Christos Faloutsos}, {and} \bibinfo{person}{Yizhou Sun}.} \bibinfo{year}{2023}\natexlab{b}.
\newblock \bibinfo{title}{PaGE-Link: Path-based Graph Neural Network Explanation for Heterogeneous Link Prediction}.
\newblock
\showeprint[arxiv]{2302.12465}~[cs.LG]
\urldef\tempurl%
\url{https://arxiv.org/abs/2302.12465}
\showURL{%
\tempurl}


\end{thebibliography}
